\newif\ifarxiv\arxivtrue

\newif\ifnotarxiv
\ifarxiv \notarxivfalse \else \notarxivtrue \fi

\ifarxiv
    \newcommand{\arxiv}[1]{#1}
\else
    \newcommand{\arxiv}[1]{}
\fi

\ifarxiv
    \newcommand{\notarxiv}[1]{}
\else
    \newcommand{\notarxiv}[1]{#1}
\fi

\ifarxiv
    \documentclass[11pt]{article}
\else 
    \documentclass{article}
    \usepackage{ijcai23}
    
    \usepackage{times}
    \usepackage{soul}
    \usepackage{url}
    \usepackage[hidelinks]{hyperref}
    \usepackage[utf8]{inputenc}
    \usepackage[small]{caption}
    \usepackage{graphicx}
    \usepackage{amsmath}
    \usepackage{amsthm}
    \usepackage{booktabs}
    \usepackage[switch]{lineno}
    \usepackage{todonotes}
\fi

\arxiv{\usepackage[authoryear]{natbib}}
\arxiv{\usepackage[margin=1in]{geometry}}

\ifnotarxiv
    
    \newtheorem{theorem}{Theorem}
\fi

\ifnotarxiv
\fi

\title{\algname: Escaping from Moderately Constrained Saddles}

\ifarxiv
    \author{
        Dmitrii Avdiukhin \\
        Indiana University \\
        \texttt{davdyukh@iu.edu}
        \and
        Grigory Yaroslavtsev \\
        George Mason University \\
        \texttt{grigory@grigory.us} \\
    }
\else
    \author{%
        Dmitrii Avdiukhin \\
        Department of Computer Science\\
        Indiana University\\
        Bloomington, IN 47405 \\
        \texttt{davdyukh@iu.edu} \\
        \And
        Grigory Yaroslavtsev \\
        Department of Computer Science \\
        George Mason University \\
        Fairfax, VA 22030 \\
        \texttt{grigory@grigory.us} \\
    }
\fi

\ifarxiv
    \usepackage[utf8]{inputenc} 
    \usepackage[T1]{fontenc}    
    \usepackage{hyperref}       
    \usepackage{booktabs}       
    \usepackage{nicefrac}       
    \usepackage{microtype}      
    \ifarxiv
        \usepackage[table]{xcolor}
    \else
        \usepackage{xcolor}         
    \fi
    \usepackage{amsmath,amssymb,amsthm,amsfonts}
    \usepackage{todonotes}
    \usepackage[vlined,ruled,linesnumbered]{algorithm2e}
    \usepackage{paralist}
    \usepackage{verbatim}
    \usepackage{framed} 
    \usepackage{etoolbox}
    \usepackage[framemethod=tikz]{mdframed}

    \usepackage{color}
    \definecolor{cornellred}{rgb}{0.7, 0.11, 0.11}
    \definecolor{dgreen}{rgb}{0.0, 0.5, 0.0}
    \definecolor{ballblue}{rgb}{0.13, 0.67, 0.8}
    \definecolor{royalblue(web)}{rgb}{0.25, 0.41, 0.88}
    \definecolor{bleudefrance}{rgb}{0.19, 0.55, 0.91}
    \definecolor{royalazure}{rgb}{0.0, 0.22, 0.66}
    \hypersetup{
        colorlinks = true,
        linkcolor = cornellred,
        citecolor = royalazure,
        linkbordercolor = white,
        urlcolor  = cornellred,
        breaklinks=true
    }
\fi
\usepackage{url}

\usepackage{graphicx,subcaption}
\usepackage{pgfplots}
\pgfplotsset{compat=newest}

\let\oldnl\nl
\newcommand{\nonl}{\renewcommand{\nl}{\let\nl\oldnl}}

\newcommand{\R}{\mathbb R}
\newcommand{\E}{\mathbb E}
\newcommand{\eps}{\varepsilon}
\newcommand{\ball}{\mathcal B}
\newcommand{\InnerProd}[1]{\langle #1 \rangle}
\newcommand{\boundary}[1]{\partial #1}
\newcommand{\interior}[1]{\Int #1}

\newcommand{\vb}{\mathbf b}

\newcommand{\ve}{\mathbf e}

\newcommand{\vg}{\mathbf g}
\newcommand{\vh}{\mathbf h}

\newcommand{\vm}{\mathbf m}

\newcommand{\vp}{\mathbf p}

\newcommand{\vu}{\mathbf u}
\newcommand{\vv}{\mathbf v}
\newcommand{\vw}{\mathbf w}
\newcommand{\vx}{\mathbf x}
\newcommand{\vy}{\mathbf y}

\newcommand{\vzero}{\mathbf 0}

\newcommand{\vA}{\mathbf A}
\newcommand{\vB}{\mathbf B}

\newcommand{\vH}{\mathbf H}
\newcommand{\vM}{\mathbf M}
\newcommand{\vO}{\mathbf O}
\newcommand{\vQ}{\mathbf Q}
\newcommand{\vP}{\mathbf P}
\newcommand{\vLambda}{\mathbf \Lambda}

\newcommand{\linecomment}[1]{\texttt{\color{blue}// #1}}

\ifarxiv
    \newcommand{\defthm}[1]{\newmdtheoremenv[roundcorner=10, innerleftmargin=7, innerrightmargin=7, leftmargin=-7, rightmargin=-7, backgroundcolor=#1!1.5, nobreak=true]}
\else
    \newcommand{\defthm}[1]{\newtheorem}
\fi

\defthm{blue}{theorem}{Theorem}[section]

\newtheorem{remark}[theorem]{Remark}

\defthm{yellow}{lemma}[theorem]{Lemma}
\defthm{yellow}{fact}[theorem]{Fact}
\defthm{green}{definition}[theorem]{Definition}
\defthm{red}{assumption}{Assumption}

\DeclareMathOperator*{\argmin}{argmin}
\DeclareMathOperator*{\rank}{rank}

\DeclareMathOperator*{\diag}{diag}
\DeclareMathOperator*{\Int}{Int}
\DeclareMathOperator*{\proj}{Proj}
\DeclareMathOperator*{\vrsg}{VRSG}
\DeclareMathOperator*{\poly}{poly}

\newcommand{\lmax}{\lambda_{|max|}}
\newcommand{\lmin}{\lambda_{\min}}
\newcommand{\grad}{\vg}
\newcommand{\step}{\eta}

\newcommand{\mcAlg}{\textsc{HoudiniEscape}}

\newcommand{\act}{\mathcal A}
\newcommand{\actInd}{\mathcal I}

\newcommand{\ourNoise}{\xi}
\newcommand{\trueR}{\sqrt[3]{\nicefrac{\delta}{\rho}}}

\newcommand{\s}[1]{#1_\perp}
\newcommand{\algname}{HOUDINI}

\newlength{\nummargin} 
\makeatletter
\patchcmd{\algocf@printnl}{\kern\skiplinenumber}{\kern\dimexpr\skiplinenumber +\nummargin - 1ex }{}{}
\makeatother

\newenvironment{algbox}
{\parindent-\nummargin\begin{framed}}
{\end{framed}}

\begin{document}

\maketitle

\begin{abstract}
    We give polynomial time algorithms for escaping from high-dimensional saddle points under a
    moderate number of constraints. Given gradient access to a smooth function $f \colon \mathbb R^d \to \mathbb R$
    we show that (noisy) gradient descent methods can escape from saddle points under a logarithmic number of
    inequality constraints.
    This constitutes progress (without reliance on NP-oracles or altering the definitions to only account for certain constraints) on the main open question of the breakthrough work of \ifarxiv~\citet{GHJY15} \else Ge et al.~\cite{GHJY15} \fi who showed an analogous result for unconstrained and equality-constrained problems. Our results hold for both regular and stochastic gradient descent.
\end{abstract}

\section{Introduction}
\label{sec:intro}

Achieving convergence of gradient descent to an (approximate) local minimum is a central question in non-convex optimization for machine learning. In the recent years, breakthrough progress starting with the work of~\citet{GHJY15} has led to a
flurry of results in this area (see e.g.~\citet{JGNKJ17,DJLJSP17,MOJ18,JNJ18,CDHS17,CD18,SRKKS19,CD20,JNGKJ21}), culminating
in almost optimal bounds~\citep{ZL21}. However, despite this success a key open question of~\citep{GHJY15} still
remains unanswered -- can gradient methods efficiently escape from saddle points in \emph{constrained} non-convex
optimization? In fact, even basic linear inequality constraints still remain an obstacle: ``Dealing with
inequality constraints is left as future work''~\citet{GHJY15}\footnote{Using Lagrangian multipliers, equality constraints can be seen as reducing the dimension of the
otherwise unconstrained problem.}. This is due to NP-hardness of the related copositivity problem~\citep{MK85}, which corresponds to the case when the number of constraints is linear in the dimension.
In this paper we make progress on this open question in the case when the number of constraints depends moderately on the dimension.

Consider a feasible set defined by $k$ linear inequality constraints: 
$S = \{\vx \in \R^d \mid \vA \vx \le \vb\},$
where $\vA \in \R^{k \times d}$ and $\vb \in \R^k$.
Let $\ball_d(\vx, r)$ be a $d$-dimensional closed ball of radius $r$ centred at $\vx$. We write $\ball(\vx, r)$ when the dimension is clear from the context and drop the first parameter when $\vx=\vzero$.
Our goal is to minimize the objective function $f \colon \R^d \to \R$ over $S$, i.e. $\min_{x \in S} f(x)$. We first introduce standard smoothness assumption.

\begin{assumption}[Smoothness]
    \label{ass:lipschitz}
    The objective function $f$ satisfies the following properties:
    \begin{compactenum}
        \item (First order) $f$ has an $L$-Lipschitz gradient ($f$ is $L$-smooth): $\|\nabla f(\vx) - \nabla f(\vy)\| \le L
        \|\vx - \vy\|$,  $\forall \vx, \vy \in \R^d$.
        \item (Second order) $f$ has a $\rho$-Lipschitz Hessian: $\|\nabla^2 f(\vx) - \nabla^2 f(\vy)\| \le \rho \|\vx - \vy\|, \forall \vx, \vy \in \R^d$.
    \end{compactenum}
\end{assumption}

%
%



\begin{definition}[Local minimum]
    For $S \subseteq \R^d$, let $f:\ \R^d \to \R$.
    A point $\vx^\star$ is a local minimum of $f$ in $S$ if and only if there exists $r > 0$ such that
    $f(\vx) \ge f(\vx^\star)$ for all $\vx \in S \cap \ball(\vx^\star, r)$.
\end{definition}
Since finding a local minimum is NP-hard even in the unconstrained case (see e.g.~\citet{AG16} and the references within) the notion of a local minimum is typically relaxed as follows.
\begin{definition}[Approximate local minimum]
    \label{def:approx_local_min}
    For $S \subseteq \R^d$ and $f:\ \R^d \to \R$ a point $\vx^\star$ is a $(\delta,r)$-approximate local minimum if
    $f(\vx) \ge f(\vx^\star) - \delta$ for all $\vx \in S \cap \ball(\vx^\star, r)$.
\end{definition}

For smooth functions one can define stationary points in terms of the gradient and the eigenvalues instead:
\begin{definition}[\citep{NP06,JNGKJ21}]
    \label{def:jin_sosp}
    A point $\vx$ is an $\eps$-second-order stationary point ($\eps$-SOSP) if $\|\nabla f(\vx)\| < \eps$ and $\lmin(\nabla^2 f(\vx)) > -\sqrt{\rho \eps}$, where $\lmin$ denotes the smallest eigenvalue.
\end{definition}


When applying this definition to the constrained case, eigenvectors and eigenvalues are not well-defined
    since there might be no eigenvectors inside the feasible set, while an escaping direction might exist.
Moreover, for $f(\vx) = -\frac 12 \|\vx\|^2$ and any compact feasible set, the Hessian is $-I$ at any point with $\lmin(-I) = -1$. Hence an $\eps$-SOSP doesn't exist according to the Definition~\ref{def:jin_sosp}, even though a local minimum exists.
In fact, Definition~\ref{def:jin_sosp} arises from the Taylor expansion, which justifies the choice of $\sqrt{\rho\epsilon}$ as the bound on the smallest eigenvalue.
If the function has a $\rho$-Lipschitz Hessian:
\[
    \left|f(\vx + \vh) - f(\vx) - \vh^\top \nabla f(\vx) - \frac 12 \vh^\top \nabla^2 f(\vx) \vh \right|
    \le \frac \rho 6  \|\vh\|^3
.\]
To guarantee that the discrepancy between the function and its quadratic approximation is small relative to $\delta$ (from Definition~\ref{def:approx_local_min}), a natural choice of $r$ is $\sqrt[3]{\nicefrac{\delta}{\rho}}$, which bounds the discrepancy with $\Theta(\delta)$.
Therefore, based on the quadratic approximation, one can distinguish
    a $(\delta,r)$-approximate local minimum from not a $(c\delta,r)$-approximate local minimum for $c < 1$.
By setting this $r$ and selecting $\eps = \sqrt[3]{\delta^2 \rho}$, we have $\sqrt{\rho \eps} = \sqrt[3]{\delta \rho^2}$ and for any $\vh \in \ball(\vx, r)$ (see Appendix~\ref{app:facts}):
\[
    f(\vx + \vh) - f(\vx)
    \ge \vh^\top \nabla f(\vx) + \frac 12 \vh^\top \nabla^2 f(\vx) \vh - \frac{\rho}{6} \|\vh\|^3
    \ge - 2 \delta
\]
%
Using the ball radius discussed above we arrive at the following version of Definition~\ref{def:approx_local_min}:

\begin{definition}[Approximate SOSP]
    \label{def:sosp}
    For $S \subseteq \R^d$, let $f:\ \R^d \to \R$ be a twice-differentiable function with a $\rho$-Lipschitz Hessian.
    A point $\vx^\star$ is a $\delta$-second-order stationary point ($\delta$-SOSP) if for $r = \sqrt[3]{\nicefrac{\delta}{\rho}}$:
    \[\inf_{\vx \in S \cap \ball(\vx^\star, r)} f(\vx) \ge f(\vx^\star) - \delta\]
\end{definition}
Note that, while Definition~\ref{def:sosp} doesn't seemingly use second-order information, our choice of the ball radius guarantees that the function is close to its quadratic approximation.
In particular, we can determine whether $f$ is a $c\delta$-SOSP for a constant $c$ by only using second-order information.

\subsection{Our Contribution}

Our results hold for stochastic gradient descent (SGD):

\begin{assumption}
    \label{ass:sgd}
    Access to a \emph{stochastic gradient oracle} $\grad(\vx)$:
    \begin{compactenum}
        \item (Unbiased expectation) $\E[\grad(\vx)] = \nabla f(\vx)$.
        \item (Variance) $\E [\|\grad(\vx) - \nabla f(\vx)\|^2] \le \sigma^2$.
    \end{compactenum}
\end{assumption}

Our main result is the following theorem which quantifies the complexity of finding an approximate SOSP under a moderate number of linear inequality constraints, showing that this problem is solvable in polynomial time for $k = O(\log d)$.
We refer to a function as $(L,\rho)$-smooth if it satisfies Assumption~\ref{ass:lipschitz} and simply \emph{second-order smooth} if both smoothness parameters are constant.

\begin{theorem}
    \label{thm_general_case_sgd}
    Let $S$ be a set defined by an intersection of $k$ linear inequality constraints.
    Let $f$ be a second-order smooth bounded function.
    Given access to a stochastic gradient oracle satisfying Assumption~\ref{ass:sgd},
        there exists an algorithm which for any $\delta > 0$ finds a $\delta$-SOSP
        in $\tilde O(\frac{1}{\delta} (d^3 2^k + \frac{d^2 \sigma^2}{\delta^{\nicefrac{4}{3}}}))$ time
        using $\tilde O(\frac{d}{\delta} (1 + \frac{d \sigma^2}{\delta^{\nicefrac{4}{3}}}))$ stochastic gradient oracle calls.
In the deterministic gradient case ($\sigma=0$), the time complexity is $\tilde O(\frac{d^3 2^k}{\delta})$ and the number of gradient oracle calls is $\tilde O(\frac{d}{\delta})$.
\end{theorem}


The exponential dependence of time complexity on $k$ in our results (not required in the oracle calls) is most likely unavoidable due to the following hardness result, which implies that when $k = d$ then the complexity of this problem can't be polynomial in $d$ under the standard hardness assumptions.
\begin{remark}[Matrix copositivity~\citep{MK85}]\label{rem:copositivity}
For a quadratic function $f(\vx) = \vx^T\vM\vx$ subject to constraints $x_i \ge 0$ for all $i$, it is NP-hard to decide whether there exists a solution with $f(\vx) > 0$.
\end{remark}


Related results in convex optimization are covered in~\citet{BV04,B15}. Among related results in non-convex optimization here we only focus on the algorithms which only use gradient information.

\subsection{Related Work}

\paragraph{Unconstrained optimization}
Recall that an \emph{$\epsilon$-first-order stationary point} ($\epsilon$-FOSP) is defined so that $\|\nabla f(\vx)\| \le \epsilon$.
Analyses of convergence to an $\eps$-FOSP are a cornerstone of non-convex optimization (see e.g. classic texts~\citet{B97,NW99}).
Quantitative analysis of convergence to an $\epsilon$-SOSP (Definition~\ref{def:jin_sosp}) started with the breakthrough work by~\citet{GHJY15} further refined in~\citet{JGNKJ17,CD18,JNJ18,CD20,JNGKJ21} and most recently in~\citet{ZL21}, who show an almost optimal bound. 
Due to the prevalence of SGD in deep learning, stochastic methods have attracted the most attention (see~\citet{A18,AL18,FLLZ18,TSJRJ18,XRY18,ZG20,ZXG20} for the case of Lipschitz gradients and~\citet{GHJY15,DKLH18} for non-Lipschitz gradients).
For an in-depth summary of the previous work on unconstrained non-convex optimization we refer the reader to~\citet{JNGKJ21}.

\paragraph{Constrained optimization}
The case of equality constraints is typically reducible to the unconstrained case by using Lagrangian multipliers (see e.g.~\citet{GHJY15}).
However, the general constrained case is substantially more challenging since even the definitions of stationarity require a substantial revision.
For first-order convergence a rich literature exists, covering projected gradient, Frank-Wolfe, cubic regularization, etc (see e.g.~\citet{MOJ18} and the references within).
For second-order convergence the landscape of existing work is substantially sparser due to NP-hardness (Remark~\ref{rem:copositivity},~\citep{MK85}).
A large body of work focuses on achieving convergence using various forms of NP-oracles (see e.g.~\citet{BCY15,CGT18,MOJ18,HLY19,NR20}), while another approach is to define stationarity in terms of tight constraints only~\citep{AJY19,LRYHH20}.

\paragraph{Relationship with other definitions of SOSP} 
As discussed in Remark~\ref{rem:copositivity}, second-order constrained optimization is NP-hard due to the hardness of the matrix copositivity problem. Definitions of constrained SOSP in the previous work fall into two categories: 1) definitions of scaled stationary points, 1) definitions which only consider active constraints (``active constraints only'' definitions), 2) definitions that preserve the NP-hardness of the problem and rely on NP-oracles to achieve polynomial-time convergence:

\begin{enumerate}
\item (Scaled) For the constraints $\vx \ge 0$,~\citet{bian2015complexity,NW20} consider the definition of scaled SOSP.
The idea is to scale $i$-th coordinate by $x_i$: i.e. instead of bounding $\nabla f(\vx)$ it bounds $X \nabla f(\vx)$, where $X = \diag(x_1, \ldots, x_d)$, and instead of eigenvalues $\nabla^2 f(\vx)$ it considers eigenvalues of $X \nabla^2 f(\vx) X$.
For this definition, $\vx \ge 0$ restricts possible applications.
\item (``Active constraints only'') In~\citet{AJY19,LRYHH20} definitions analogous to Definition~\ref{def:jin_sosp}, and the second-order conditions are given with respect to the set of \emph{active} (i.e. tight for the current iterate) constraints. This allows to bypass the NP-hardness since the point at which the hardness of the copositivity problem applies now becomes a stationary point by definition. 
\item (NP-hard) In the results relying on NP-oracles (e.g.~\citep{BCY15,MOJ18,HLY19,NR20}) the complexity is shifted on solving black-box quadratic optimization problems of a certain type. A key advantage of these types of approaches is that they can handle an arbitrary number of constraints and hence promising in certain machine learning applications.
\end{enumerate}

What is currently lacking in the state of the art is a quantitative analysis of the complexity of convergence to a second-order stationary point, which shows full dependence on both the dimension and accuracy while defining stationarity with respect to the full set of constraints, instead of just active constraints only\footnote{While unrefereed manuscripts (\cite{HS18} and \cite{NLR20}, relying on~\cite{HS18} as a subroutine) do contain related results, our work differs in a number of important aspects. \cite{HS18} and~\cite{NLR20} require access to the exact gradient and Hessian. We only assume access to the stochastic gradient oracle. Furthermore, ~\cite{HS18} assumes access to matrix diagonalization. However, the diagonalization can only be found approximately, which compromises the stability of this approach, especially with respect to the linear term transformations. Our approach handles this issue by appropriate perturbation and using known results for matrix diagonalization.
Finally, compared with~\cite{HS18}, our main contribution is focused on solving a substantially different problem. While~\cite{HS18} find a global minimum of a quadratic problem, we find an approximate local minimum of an arbitrary smooth non-convex function.}. Our goal in Theorem~\ref{thm_general_case_sgd} is to address this gap and give such analysis. 

\subsection{Technical overview}

\newcommand{\CubeLength}{1}
\newcommand{\CubeHeight}{\CubeLength * \CubeLength * 2}
\ifarxiv
    \newcommand{\figwidth}{0.45\textwidth}
    \newcommand{\figscale}{0.7}
\else
    \newcommand{\figwidth}{0.22\textwidth}
    \newcommand{\figscale}{0.5}
\fi
\newcommand{\figlinethick}{3pt}
\newcommand{\nsamples}{10}

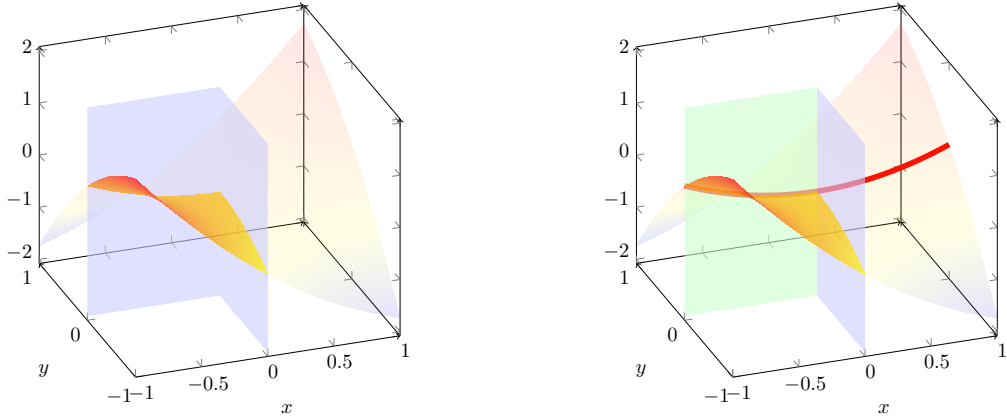
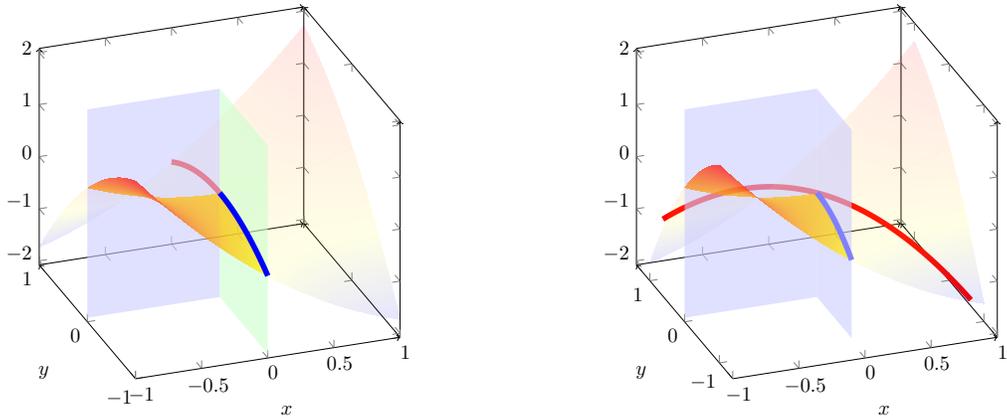
\begin{figure}[ht!]
    \centering
    \begin{subfigure}[t]{\figwidth}
        \centering
        \begin{tikzpicture}[scale=\figscale]
        \begin{axis}[
            samples=\nsamples,
            view={-20}{39.9},
            axis line style={->},
            xlabel=$x$,
            ylabel=$y$,
            z post scale=1.5
        ]
        \tikzset{declare function={%
        fx(\x,\y)=\x * 0.866025 + \y * 0.5;
        fy(\x,\y)=-\x * 0.5 + \y * 0.866025;
        }}
        \addplot3 [surf,shader=interp,draw=black,domain=-\CubeLength:0,domain y=0:\CubeLength,opacity=0.1] {fx(x,y)^2-fy(x,y)^2};
        \addplot3 [surf,shader=interp,draw=black,domain=0:\CubeLength,domain y=-\CubeLength:0,opacity=0.1] {fx(x,y)^2-fy(x,y)^2};
        \addplot3 [surf,shader=interp,draw=black,domain=0:\CubeLength,domain y=0:\CubeLength,opacity=0.1] {fx(x,y)^2-fy(x,y)^2};
    
        \fill[color=blue!20,opacity=0.6] (0,-\CubeLength,-\CubeHeight) -- (0,0,-\CubeHeight) -- (0,0,\CubeHeight) -- (0,-\CubeLength,\CubeHeight) -- cycle;
        \fill[color=blue!20,opacity=0.6] (-\CubeLength,0,-\CubeHeight) -- (0,0,-\CubeHeight) -- (0,0,\CubeHeight) -- (-\CubeLength,0,\CubeHeight) -- cycle;
    
        \addplot3 [surf,shader=interp,draw=black,domain=-\CubeLength:0,domain y=-\CubeLength:0,opacity=0.7] {fx(x,y)^2-fy(x,y)^2};
        \end{axis}
        \end{tikzpicture}
        \caption{Feasible set with two inequalities: $x \le 0, y \le 0$.}
    \end{subfigure}
    \quad
    \begin{subfigure}[t]{\figwidth}
        \centering
        \begin{tikzpicture}[scale=\figscale]
        \begin{axis}[
            samples=\nsamples,
            view={-20}{40},
            axis line style={->},
            xlabel=$x$,
            ylabel=$y$,
            z post scale=1.5
        ]
        \tikzset{declare function={%
        fx(\x,\y)=\x * 0.866025 + \y * 0.5;
        fy(\x,\y)=-\x * 0.5 + \y * 0.866025;
        }}
        \addplot3 [draw=red,line width=\figlinethick, domain=-\CubeLength:0,samples=30,samples y=1,smooth,opacity=1] ({x}, {0}, {fx(x,0)^2-fy(x,0)^2});
        \addplot3 [draw=red,line width=\figlinethick, domain=0:\CubeLength,samples=30,samples y=1,smooth,opacity=1] ({x}, {0}, {fx(x,0)^2-fy(x,0)^2});

        \addplot3 [surf,shader=interp,draw=black,domain=-\CubeLength:0,domain y=0:\CubeLength,opacity=0.1] {fx(x,y)^2-fy(x,y)^2};
        \addplot3 [surf,shader=interp,draw=black,domain=0:\CubeLength,domain y=-\CubeLength:0,opacity=0.1] {fx(x,y)^2-fy(x,y)^2};
        \addplot3 [surf,shader=interp,draw=black,domain=0:\CubeLength,domain y=0:\CubeLength,opacity=0.1] {fx(x,y)^2-fy(x,y)^2};
    
        \fill[color=blue!20,opacity=0.6] (0,-\CubeLength,-\CubeHeight) -- (0,0,-\CubeHeight) -- (0,0,\CubeHeight) -- (0,-\CubeLength,\CubeHeight) -- cycle;
        \fill[color=green!20,opacity=0.6] (-\CubeLength,0,-\CubeHeight) -- (0,0,-\CubeHeight) -- (0,0,\CubeHeight) -- (-\CubeLength,0,\CubeHeight) -- cycle;
    
        \addplot3 [surf,shader=interp,draw=black,domain=-\CubeLength:0,domain y=-\CubeLength:0,opacity=0.7] {fx(x,y)^2-fy(x,y)^2};
        \end{axis}
        \end{tikzpicture}
        \caption{Active constraint $y=0$ (green) with no escape direction.}
    \end{subfigure}
    
    \vspace{10pt}
    \begin{subfigure}[t]{\figwidth}
        \centering
        \begin{tikzpicture}[scale=\figscale]
        \begin{axis}[
            samples=\nsamples,
            view={-20}{40},
            axis line style={->},
            xlabel=$x$,
            ylabel=$y$,
            z post scale=1.5
        ]
        \tikzset{declare function={%
        fx(\x,\y)=\x * 0.866025 + \y * 0.5;
        fy(\x,\y)=-\x * 0.5 + \y * 0.866025;
        }}
        \addplot3 [draw=red,line width=\figlinethick,domain=0:\CubeLength,samples=30,samples y=1,smooth,opacity=1] ({0}, {x}, {fx(0,x)^2-fy(0,x)^2});

        \addplot3 [surf,shader=interp,draw=black,domain=-\CubeLength:0,domain y=0:\CubeLength,opacity=0.1] {fx(x,y)^2-fy(x,y)^2};
        \addplot3 [surf,shader=interp,draw=black,domain=0:\CubeLength,domain y=-\CubeLength:0,opacity=0.1] {fx(x,y)^2-fy(x,y)^2};
        \addplot3 [surf,shader=interp,draw=black,domain=0:\CubeLength,domain y=0:\CubeLength,opacity=0.1] {fx(x,y)^2-fy(x,y)^2};

        \fill[color=green!20,opacity=0.6] (0,-\CubeLength,-\CubeHeight) -- (0,0,-\CubeHeight) -- (0,0,\CubeHeight) -- (0,-\CubeLength,\CubeHeight) -- cycle;
        \fill[color=blue!20,opacity=0.6] (-\CubeLength,0,-\CubeHeight) -- (0,0,-\CubeHeight) -- (0,0,\CubeHeight) -- (-\CubeLength,0,\CubeHeight) -- cycle;
    
        \addplot3 [surf,shader=interp,draw=black,domain=-\CubeLength:0,domain y=-\CubeLength:0,opacity=0.7] {fx(x,y)^2-fy(x,y)^2};

        \addplot3 [draw=blue,line width=\figlinethick, domain=-\CubeLength:0,samples=30,samples y=1,smooth,opacity=1] ({0}, {x}, {fx(0,x)^2-fy(0,x)^2});
        \end{axis}
        \end{tikzpicture}
        \caption{Active constraint $x=0$ (green) with an escape direction $(0, -1)$. As shown in Lemma~\ref{lemma:copositivity_approx_eigen}, out of two escape directions $(0, -1)$ and $(0,1)$ in this constraint, at least one $(0, -1)$ (blue) lies in the feasible set, and is found by the algorithm.}
        \label{fig:escape_dir}
    \end{subfigure}
    \quad
    \begin{subfigure}[t]{\figwidth}
        \centering
        \begin{tikzpicture}[scale=\figscale]
        \begin{axis}[
            samples=\nsamples,
            view={-20}{40},
            axis line style={->},
            xlabel=$x$,
            ylabel=$y$,
            z post scale=1.5
        ]
        \tikzset{declare function={%
        fx(\x,\y)=\x * 0.866025 + \y * 0.5;
        fy(\x,\y)=-\x * 0.5 + \y * 0.866025;
        gx(\x,\y)=\x * 0.866025 - \y * 0.5;
        gy(\x,\y)=\x * 0.5 + \y * 0.866025;
        }}
        \addplot3 [draw=red,line width=\figlinethick,domain=-1.6 * \CubeLength:0,samples=30,samples y=1,smooth,opacity=1] ({gx(0,x)}, {gy(0,x)}, {fx(0,x)^2-fy(0,x)^2});
        \addplot3 [draw=red,line width=\figlinethick,domain=0:1.6 * \CubeLength,samples=30,samples y=1,smooth,opacity=1] ({gx(0,x)}, {gy(0,x)}, {fx(0,x)^2-fy(0,x)^2});

        \addplot3 [surf,shader=interp,draw=black,domain=-\CubeLength:0,domain y=0:\CubeLength,opacity=0.1] {fx(x,y)^2-fy(x,y)^2};
        \addplot3 [surf,shader=interp,draw=black,domain=0:\CubeLength,domain y=-\CubeLength:0,opacity=0.1] {fx(x,y)^2-fy(x,y)^2};
        \addplot3 [surf,shader=interp,draw=black,domain=0:\CubeLength,domain y=0:\CubeLength,opacity=0.1] {fx(x,y)^2-fy(x,y)^2};
        \fill[color=blue!20,opacity=0.6] (0,-\CubeLength,-\CubeHeight) -- (0,0,-\CubeHeight) -- (0,0,\CubeHeight) -- (0,-\CubeLength,\CubeHeight) -- cycle;
    
        \fill[color=blue!20,opacity=0.6] (-\CubeLength,0,-\CubeHeight) -- (0,0,-\CubeHeight) -- (0,0,\CubeHeight) -- (-\CubeLength,0,\CubeHeight) -- cycle;
    
        \addplot3 [surf,shader=interp,draw=black,domain=-\CubeLength:0,domain y=-\CubeLength:0,opacity=0.7] {fx(x,y)^2-fy(x,y)^2};
        \addplot3 [draw=blue!50,line width=\figlinethick, domain=-\CubeLength:0,samples=30,samples y=1,smooth,opacity=1] ({0}, {x}, {fx(0,x)^2-fy(0,x)^2});

        \end{axis}
        \end{tikzpicture}
        \caption{When no constraints are active, the algorithm finds red directions (negative eigenvectors) outside the feasible set. Note that escape directions with no active constraints exist (blue). Lemma~\ref{lemma:copositivity_approx_eigen} guarantees that we find them in some constrained space (Figure~\ref{fig:escape_dir}).}
    \end{subfigure}
    \caption{Function $f(x,y) = (\frac{\sqrt 3}{2}x + \frac 1 2 y)^2 - (-\frac 12 x + \frac{\sqrt 3}{2} y)^2$ (composition of $x^2-y^2$ and rotation by $\nicefrac{\pi}{6}$).} \label{fig:escape-illustration}
\end{figure}

We address the NP-hardness of the copositivity problem by focusing on the case of a moderate number of constraints and arguing that it can be addressed using gradient-based methods. In order to streamline the presentation we first focus on the key challenge of escaping from a saddle point in a corner defined by the constraints when the underlying function is simply quadratic (Section~\ref{sec:copositivity}). This is already enough to capture the some of the key contributions, while more technical details and the full algorithm are given in Section~\ref{sec:general-case}.

\paragraph{Quadratic corner saddle point (Section~\ref{sec:copositivity})}
In this simplified scenario the NP-hardness comes from the fact that the point we aim to find can lie in the intersection of an arbitrary subset of constraints. By doing an exhaustive search over this set of constraints (Algorithm~\ref{alg:matrix_copositivity}) and enforcing them throughout the search process we are able to reduce to a setting similar to the equality-constrained case (Algorithm~\ref{alg:eigenvector}).
We show different subsets of constraints enforced by the algorithm in an example in Figure~\ref{fig:escape-illustration}.
The key challenge is making this argument formal and arguing that this process converges to a constrained approximate SOSP as in Definition~\ref{def:sosp}.
This relies on performing a robust analysis of the properties of the points which lead to the hardness of copositivity. In particular, we show that after we guess the set of constraints correctly, the problem reduces to finding the smallest eigenvector (Lemma~\ref{lemma:copositivity_check}).
Exact error analysis of the eigenvector process (Lemma~\ref{lemma:copositivity_approx_eigen}, Lemma~\ref{lem:approx_eigenvector}) is then required to complete the proof of the main theorem (Theorem~\ref{thm:copositive_case}).

\paragraph{General case (Section~\ref{sec:general-case})}
In the algorithm for the general case we address the three assumptions made in the quadratic corner saddle point case, while also handling the stochasticity in the gradient. The latter part is standard and is handled via variance reduction in the Algorithm~\ref{alg:vrsg} (analyzed in Lemma~\ref{lem:vrsg} and Lemma~\ref{lem:stochastic-approximation}).
The full algorithm iterates the escape subroutine (Algorithm~\ref{alg:escaping_general}) until an escaping point is found.
The escape subroutine first approximates the Hessian matrix using the gradient oracle and then performs exhaustive search over the set of active constraints at the escaping point in a way similar to the quadratic corner case. After the correct subset of constraints is fixed the current iterate needs to be projected on this set of constraints, which also necessitates a recomputation of various related parameters. When this is done the problem is solved by a subroutine Algorithm~\ref{alg:eigenvector_general}, analogous to Algorithm~\ref{alg:eigenvector} from the quadratic corner case.

However, since the function is no longer quadratic and the gradient can be large several modifications are required. The algorithm tries to find an escaping direction within a ball of radius $r$, within which the function is well approximated by a quadratic by the smoothness assumption (as discussed above).
First, the algorithm tries to escape using the gradient term.
If that doesn't work then we consider two cases: 1) the solution is inside the ball of radius $r = \trueR$ (from Definition~\ref{def:sosp}), 2) the solution is on the boundary of this ball. In the first case the solution is a critical point and hence can be found using the Newton step. In the second case we first diagonalize the matrix using an orthogonal transformation.
This gives a quadratic function with a linear term whose critical points on the boundary of the ball can be found explicitly (up to required precision) as roots of the corresponding polynomial.
Here we note that the diagonalization performed in this case is the most computationally expensive step in the algorithm, resulting in polynomial dependence on the dimension.

\paragraph{Notation}

For a set $S$ let $\interior{S}$ be its interior and $\boundary{S}$ be its boundary.
For $\vx \in \R^d$ and $S \subseteq \R^d$, $\proj_S(\vx) = \argmin_{\vy \in S}\|\vx - \vy\|$ is the projection of $\vx$ on $S$.
For a square matrix $\vM$ with eigenvalues $\lambda_1 \le \ldots \le \lambda_d$, we denote $\lmin(\vM) = \lambda_1$ and $\lmax(\vM) = \max (|\lambda_1|, |\lambda_d|)$.
For $S = \{\vx \in \R^d \mid \vA \vx \le \vb\}$ and $\vx \in S$, we say that $i$-th constraint is active at $\vx$ if $\vA_i^\top \vx = b_i$, where $\vA_i$ is the $i$-th row of $\vA$.
$\tilde O$ notation hides polylogarithmic dependence on all parameters, including error probability.

\ifarxiv
    For the full list of notation refer to Table~\ref{tab:notation_full}.
    \begin{table*}[!t]
    \centering
    \caption{Notation used in the paper.}
    \label{tab:notation_full}
    \begin{tabular}{ll}
    \hline
         Notation & Explanation \\
    \hline
    \hline
        \multicolumn{2}{l}{\textbf{General notation}} \\
    \hline
        $d$ & Dimension \\
        $k$ & Number of linear inequality constraints \\
        $S \subseteq \R^d$ & Feasible set, defined by $k$ linear inequality constraints: $S = \{\vx \mid \vA \vx \le \vb\}$  \\
        $\lmax(\vM)$ & For matrix $\vM$ with eigenvalues $\lambda_1\le \ldots \le \lambda_d$, $\lmax(\vM)$ is $\max_i |\lambda_i|$ \\
        $\lmin(\vM)$ & For matrix $\vM$ with eigenvalues $\lambda_1\le \ldots \le \lambda_d$, $\lmin(\vM)$ is $\lambda_1$ \\
        $\actInd$ & Subset of constraints (usually the enforced set of active constraints): $\actInd \subseteq [k]$ \\
        $\act$ & Affine subspace of $\R^d$ (usually defined by enforcing active constraints from $\actInd$) \\
        $\proj_{\act}(\vx)$ & Projection of $\vx$ on $\act$: \quad $\argmin_{\vy \in \act} \|\vy - \vx\|$ \\
        $\ball_d(\vx, r)$ & Closed $d$-dimensional ball of radius $r$: \ $\ball_d(\vx, r) = \{\vy \in \R^d \mid \|\vx - \vy\| \le r\}$ \\
        $\ball(r)$ & Closed zero-centered ball of radius $r$: \ $\ball(r) = \{\vy \mid \|\vy\| \le r\}$ \\
        $\interior{S}$ & Interior of a set $S$ \\
        $\boundary{S}$ & Boundary of a set $S$ \\
        $\tilde O$ & Hides polylogarithmic dependence on all parameters, including error probability \\
    \hline
    \hline
        \multicolumn{2}{l}{\textbf{Problem parameters}} \\
    \hline
        $f:\ \R^d \to \R$ & Objective function \\
        $L$ & Lipschitz constant of $\nabla f$: \quad $\|\nabla f(x) - \nabla f(y)\| \le L \|x - y\|$ \\
        $\rho$ & Lipschitz constant of $\nabla^2 f$: \quad $\|\nabla^2 f(x) - \nabla^2 f(y)\| \le \rho \|x - y\|$ \\
        $\sigma^2$ & Variance of stochastic gradient $\vg$:\quad $\E[\|\vg(\vx) - \nabla f(\vx)\|^2] \le \sigma^2$ \\
        $r$ & Optimization radius: for a saddle point $\vx$, we are looking for $\min_{\vy \in \ball(\vx, r) \cap S} f(\vy)$ \\
        $\delta$ & For a saddle point, possible function improvement in $\ball(\vx, r) \cap S$ (Definition~\ref{def:sosp}) \\
    \hline
    \hline
        \multicolumn{2}{l}{\textbf{Notation from Section~\ref{sec:copositivity}}} \\
    \hline
        $\vM$ & Matrix defining $f$:\quad $f = \frac 12 \vx^\top \vM \vx$ \\
        $\act_\actInd$ & Affine subspace defined by constraints from $\actInd$: $\act_\actInd = \{\vx \mid \vA_i^\top \vx = b_i \text{ for all } i \in \actInd\}$ \\
    \hline
    \hline
        \multicolumn{2}{l}{\textbf{Notation from Section~\ref{sec:general-case}}} \\
    \hline
        $\vrsg(\vx, \tilde{\sigma})$ & Subroutine returning vector $\vg$ such that $\|\vg - \nabla f(\vx)\| < \tilde{\sigma}$ w.h.p. \\
        $\vM$ & Approximation of $\nabla^2 f(\vzero)$ + noise $\gamma I$ \\
        $\vv$ & Approximation of $\nabla f(\vzero)$ + noise $\zeta$ \\
        $\vO$ & Orthonormal basis of $\act$ \\
        $\vp$ & Projection of the saddle point on $\act$ \\
        $r=\trueR$ & Optimization radius from Definition~\ref{def:sosp} \\
        $\s{\vM}, \s{\vv}, C$ & When working in affine subspace, we optimize $g(\vx) = \frac 12 \vx^\top \s{\vM} \vx + \vx^\top \s{\vv} + C$ \\
        $\s{S}, \s{r}$ & When working in affine subspace, we optimize $g$ in $\s{S} \cap \ball(\s{r})$ \\
        $\vLambda, \vQ$ & Diagonalization of $\s{\vM}$:\quad $\s{\vM} \approx \vQ^\top \vLambda \vQ$, with diagonal $\vLambda$ and orthogonal $\vQ$ \\
        $\tilde{\vv}$ & Rotation of $\s{\vv}$:\quad $\tilde{\vv} = \vQ \s{\vv}$ \\
        $\mu$ & Multiplier from the method of Lagrangian multipliers \\
    \hline
    \end{tabular}
\end{table*}

\fi
\section{Quadratic Corner Saddle Point Case}
\label{sec:copositivity}

We introduce the key ideas of the analysis in a simplified setting when: 1) the function $f$ is quadratic, 2) the gradient is small, 3) the current iterate is located in a corner of the constraint space. Intuitively, this represents the key challenge of the constrained saddle escape problem since its NP-hardness comes from the hardness of the matrix copositivity problem in Remark~\ref{rem:copositivity} (i.e. the function is exactly quadratic and has no gradient at the current iterate which lies in the intersection of all constraints). We refer to this setting as the Quadratic Corner Saddle Point problem defined formally below.
By shifting the coordinate system, w.l.o.g. we can assume that the saddle point is $\vzero$ and $f(\vzero)=0$\footnote{Algorithms~\ref{alg:matrix_copositivity} and~\ref{alg:eigenvector} don't require saddle point $\vx$ to be $\vzero$. All the statements are trivially adapted for the case when $\vx$ is not $\vzero$}.
If $\vzero$ is a ($\delta,r$)-QCSP (as defined below), the objective can be decreased by $\delta$ within a ball of radius $r$.
\begin{definition}[Quadratic Corner Saddle Point]
    \label{def:qscp}
    Let $S = \{\vx \mid \vA \vx \le \vzero\}$.
    For $\delta, r > 0$ and function $f(\vx) = \frac 12 \vx^\top \vM \vx$, we say that a point $\vzero$ is a:
    \begin{compactitem}
        \item ($\delta, r$)-Quadratic Corner Saddle Point (($\delta,r$)-QCSP) if
            $\min_{\vx \in \ball(r) \cap S} f(\vx) < -\delta$.
        \item ($\delta,r$)-boundary QCSP if $\min_{\vx \in \ball(r) \cap \boundary{S}} f(\vx) < -\delta$.
    \end{compactitem}
\end{definition}

\begin{algorithm}
    \caption{\mcAlg\textsc{Corner}: Escaping from a corner for a quadratic function}
    \label{alg:matrix_copositivity}
    \SetKwInOut{Input}{input}
    \Input{Starting point $\vx$, feasible set $S = \{\vy \mid \vA (\vy - \vx) \le \vzero \in \R^k\}$}
    \nonl\textbf{parameters:} $\delta$ and $r$ from definition of $(\delta,r)$-QCSP
    \For{$\actInd \in 2^{[k]}$~-- every subset of constraints}{
        $\act \gets \{\vy \mid \vA_i^\top (\vy - \vx) = 0 \text{ for } i \in \actInd\}$, where \ $\vA_i$ is the $i$-th row of $\vA$ \hfill\linecomment{Optimize in $\act$} \\
        $\vy \gets \textsc{FindInsideCorner}(\vx, \act)$ \\
        \If{$\vy \in S$ and $f(\vy) < f(\vx) - \frac{\delta}{2}$}{
            \Return $\vy$
        }
    }
    \Return $\bot$
\end{algorithm}
\begin{algorithm}
    \caption{\textsc{FindInsideCorner}$(\vx, \act)$}
    \label{alg:eigenvector}
    \SetKwInOut{Input}{input}
    \Input{Corner $\vx$, affine subspace $\act$ with $\vx \in \act$}
    \nonl\textbf{parameters:} $\delta$ and $r$ from Definition~\ref{def:qscp}, step size $\step=\frac 1L$, number of iterations $T=\tilde O(\frac{L r^2}{\delta})$ \\
    Sample $\ourNoise \sim \mathcal{N}(\vzero, I)$,  
    $\vx_0 \gets \proj_{\act}(\vx + \ourNoise)$ \\
    \For{$t = 0, \ldots, T-1$}{
        \linecomment{Power method step} \\
        $\vx_{t+1} \gets \proj_{\act}(\vx_t - \step (\nabla f(\vx_t) - \nabla f(\vx)))$ \label{line:power_method} \\
    }
    $\ve \gets r \frac{\vx_{T} - \vx}{\|\vx_{T} - \vx\|}$ \\
    \Return $\vx + \ve$
\end{algorithm}

In this section, we show how to escape from a $(\delta,r)$-QCSP
(see Appendix~\ref{app:copositivity} for the full proof):
\begin{theorem}[Quadratic Corner Saddle Point Escape]
    \label{thm:copositive_case}
    Let $\delta, r > 0$.
    Let $f(\vx) = \frac 12 \vx^\top \vM \vx$ with $\lmax(\vM) \le L$ and
        let $S = \{\vx \mid \vA \vx \le \vzero\}$ be defined by $k$ linear inequalies.
    If $\vzero$ is a ($\delta,r$)-QCSP, then Algorithm~\ref{alg:matrix_copositivity} with probability at least $1-\xi$ finds a point $\vx \in S \cap \ball(r)$ with $f(\vx) < -\Omega(\delta)$
        using $O\left(\frac{L r^2 k 2^k}{\delta} \log \frac{1}{\xi}\right)$ deterministic gradient oracle calls.
\end{theorem}

For the rest of the section, we assume that $\vzero$ is a $(\delta,r)$-QCSP, i.e. $\min_{\vx \in S \cap \ball(r)} f(\vx) < -\delta$.
We consider two cases depending on whether $\vzero$ is a $(\delta,r)$-boundary QCSP.

\paragraph{Case 1: $\vzero$ is a $(\delta,r)$-boundary QCSP.}
For a subset of inequality constraints $\actInd \subseteq [k]$ we define the subspace where these constraints are active: $\act_\actInd = \{\vx \mid \vA_i^\top \vx = 0 \text{ for all } i \in \actInd\}$.
Let $\actInd$ be a maximal\footnote{As we don't know $\actInd$, Algorithm~\ref{alg:matrix_copositivity} tries all subsets of constraints.} subset of constraints such that $\min_{\vx \in \act_\actInd \cap \ball(r)} f(\vx) < - \delta$.
If $\vP$ is a projection operator on $\act_\actInd$, it suffices to optimize $g(\vx) := f(\vP \vx) = \frac 12 \vx^\top (\vP \vM \vP) \vx$.
Therefore, we reduced the original problem to minimizing a different quadratic form in the same feasible set.
For any $i \in \actInd$, $\vA_i^\top \vP \vx \le 0$ holds trivially, since $\vA_i^\top \vy = 0$ for any $\vy \in \act$, and hence constraints from $\actInd$ can be ignored.
If a constraint not from $\actInd$ is active in $\vP \vx$, then $f(\vP \vx) \ge -\delta$, since $\actInd$ is a maximal subset of constraints with $\min_{\vx \in \act_\actInd \cap \ball(r)} < - \delta$.
Therefore, this reduces case to the next case.

\paragraph{Case 2: $\vzero$ is not a $(\delta,r)$-boundary QCSP.}
In this case, we show that any $\vx \in \ball(r)$ with $f(\vx) < -\delta$ must lie in $S$, and for $f(\vx) = \frac 12 \vx^\top \vM \vx$ it suffices to find the eigenvector corresponding to the smallest eigenvalue of $\vM$.
We first show that there exists an eigenvector which improves the objective.
\begin{lemma}
    \label{lemma:copositivity_check}
    Let $f(\vx) = \frac 12 \vx^\top \vM \vx$ and $S = \{\vx \mid \vA \vx \le \vzero\}$.
    If $\vzero$ is not a $(\delta,r)$-boundary QCSP for $\delta, r > 0$, then the following statements are equivalent:
    \begin{compactenum}
        \item $\vzero$ is $(\delta,r)$-QCSP, i.e. $\min_{\vx \in S \cap \ball(r)} f(\vx) < -\delta$.
        \item There exists an eigenvector $\ve$ of $\vM$ such that $\ve \in \interior{S} \cap \boundary{\ball(r)}$ and $f(\ve) < -\delta$.
    \end{compactenum}
\end{lemma}

Finding an exact eigenvector might be impossible.
Hence, we show that finding $\vx \in \ball(r)$ with $f(\vx) < -\delta$ suffices, since either $\vx$ or $-\vx$ are in $S$.
\begin{lemma}
    \label{lemma:copositivity_approx_eigen}
    Let $f(\vx) = \frac 12 \vx^\top \vM \vx$ and $S = \{\vx \mid \vA \vx \le \vzero\}$.
    For $\delta, r > 0$ and $\hat{\vx} \in \boundary{\ball(r)}$, if $f(\hat{\vx}) < - \delta$ and the following conditions hold, then either $\hat{\vx} \in S$ or $-\hat{\vx} \in S$:
    \begin{compactenum}
        \item $\min_{\vx \in S \cap \ball(r)} f(\vx) < -\delta$, i.e. $\vzero$ is a $(\delta,r)$-QCSP,
        \item $\min_{\vx \in \boundary{S} \cap \ball(r)} f(\vx) \ge -\delta$, i.e $\vzero$ is not a $(\delta,r)$-boundary QCSP,
    \end{compactenum}
\end{lemma}
To show Lemma~\ref{lemma:copositivity_approx_eigen}, we use the fact that, by Lemma~\ref{lemma:copositivity_check},
    there exists an eigenvector $\ve$ with $\ve \in \interior{S} \cap \boundary{\ball(r)}$ and $f(\ve) < -\delta$.
For the sake of contradiction, if both $-\hat{\vx}$ and $\hat{\vx}$ are not in $S$,
    at least one of them has non-negative inner product with $\ve$.
W.l.o.g. we assume $\hat{\vx}^\top \ve \ge 0$, and we consider an arc on $\boundary{\ball(r)}$ connecting $\hat{\vx}$ and $\ve$.
Since $\ve \in S$ and $\hat{\vx} \notin \ve$, the arc intersects $\boundary{S}$ at some point.
We show that any point $\vx$ on the arc has $f(\vx) < -\delta$, and hence this also holds for the point on the boundary, contradicting the assumption that $\vzero$ is not a $(\delta,r)$-boundary QCSP, finishing the proof.

The main idea behind the algorithm is that Algorithm~\ref{alg:eigenvector} emulates the power method on matrix $I - \frac \vM L$.
Hence, for any $\eps$ it allows us to find a vector $\vx$ such that
\[
    \frac{|\vx^\top (I - \frac{\vM}{L}) \vx|}{\|\vx\|^2} \ge (1 - \eps) \lmax\left(I - \frac{\vM}{L}\right)
.\]
Since $\lmax(\vM) \le L$, all eigenvalues of $I - \frac \vM L$ are positive, and hence the power method approximates the eigenvector corresponding to the largest eigenvalue of $I - \frac \vM L$, and hence to the smallest eigenvalue of $\vM$.

We know that $f(\vx) < -\delta$, and we aim to find $\vx \in \ball(r)$ with $f(\vx) < -(1-\eps) \delta$ for a constant $\eps$.
Since there exists an eigenvector $\ve \in \boundary{\ball(r)}$ of $\vM$ with $\frac 12 \ve^\top \vM \ve < -\delta$, we have $\lmin(\vM) < \frac{-2 \delta}{\|\ve\|^2} = \frac{-2 \delta}{r^2}$, and hence the largest eigenvalue of $I - \frac{\vM}{L}$ is at least $1 - \frac{\lmin(\vM)}{L} \ge 1 + \frac{2 \delta}{L r^2}$.
Finding $\vx$ with $\frac 12 \vx^\top \vM \vx < -(1-\eps)\delta$ is equivalent to finding $\vx$ with $\vx^\top (I - \frac \vM L) \vx \ge (1 + \frac{2 (1 - \eps) \delta}{L r^2})\|\vx\|^2$, and the power method achieves this in
$O(\log d + \frac{L r^2}{\eps \delta})$ iterations (see proof of Lemma~\ref{lem:approx_eigenvector} in Appendix~\ref{app:copositivity}).
\begin{lemma}
    \label{lem:approx_eigenvector}
    Let $\delta, r > 0$, $\vx \in \R^d$ and $\eps \in (0,1)$.
    Let $f(\vx) = \frac 12 \vx^\top \vM \vx$ with $\lmax(\vM) \le L$.
    Let $\act$ be a linear subspace of $\R^d$.
    If $\min_{\vx \in \act \cap \ball(r)} f(\vx) < -\delta$,
        then Algorithm~\ref{alg:eigenvector} finds $\vx \in \act \cap \boundary{\ball(r)}$ with $f(\vx) \le -(1-\eps) \delta$
        after $T=\tilde O\left(\frac{L r^2}{\eps \delta}\right)$ iterations w.h.p.
\end{lemma}
Finally, we now prove Theorem~\ref{thm:copositive_case}.
First, by exhaustive search we guess a maximal subset of active constraints $\actInd$ such that subspace $\act_\actInd$ formed by these linear constraints has $\vx \in \ball(r) \cap S$ with $f(\vx) < \delta$.
Using Algorithm~\ref{alg:eigenvector}, we find $\vy \in \ball(r) \cap \act_\actInd$ with $f(\vy) < -(1-\eps)\delta$. Then $\vy \in S$ by Lemma~\ref{lem:approx_eigenvector}, since $\actInd$ is a maximal subset of constraints with an escape direction.

\section{Main result}
\label{sec:general-case}


\begin{algorithm}[t!]
    \caption{\mcAlg$(\vx, S, \delta)$: Escaping from a saddle point}
    \label{alg:escaping_general}
    \SetKwInOut{Input}{input}
	\SetKwInOut{Output}{output}
	\Indentp{0.1em}
    \Input{Saddle point $\vx$, $\delta$ from definition of $\delta$-SOSP, feasible set $S = \{\vy \mid \vA \vy \le \vb \in \R^k\}$}
    \Output{either reports that $\vx$ is a $\delta$-SOSP or finds $\vu \in S \cap \ball(\vx, r)$ with $f(\vu) < f(\vx) - \Omega(\delta)$}
    Construct $f'(\vx + \vh)$~-- quadratic approximation of $f(\vx + \vh)$~-- using stochastic gradient oracle calls \\
    \For{$\actInd$~-- every subset of constraints}{
        $\act \gets \{\vy \mid \vA_i^\top \vx = b_i,\ i \in \actInd\}$, where $\vA_i$ is the $i$-th row of $\vA$ \hfill\linecomment{Optimize in $\act$}\\
        Let $\vp$ be the projection of $\vx$ on $\act$ \\
        Let $\vO \in \R^{d \times \dim \act}$ be an orthonormal basis of $\act$ \\
        Define $g(\vy) := f'(\vp + \vO \vy)$ \\
        Algorithm~\ref{alg:eigenvector_general} tries to find an escape direction for $g$ \\
        \If{Algorithm~\ref{alg:eigenvector_general} finds direction $\vy$}{\textbf{return} $\vp + \vO \vy$}
    }
    Didn't find escape direction: \textbf{report} that $\vx$ is a $\delta$-SOSP 
\end{algorithm}

\begin{algorithm}[t!]
    \caption{\\\textsc{FindInside}$(\vx, \delta, (\s{\vM}, \s{\vv}), (\s{r}, \s{S}))$}
    \label{alg:eigenvector_general}
    \SetKwInOut{Input}{input}
    \SetKwInOut{Output}{output}
	\Indentp{0.1em}
    \Input{$\delta$ from definition of $\delta$-SOSP,
    $g(\vy) = \frac 12 \vy^\top \s{\vM} \vy + \vy^\top \s{\vv}$~-- objective in $\R^{\dim \act}$,\\
    $\s{S} \cap \ball(\s{r})$~-- feasible set in $\R^{\dim \act}$
    }
    \Output{Escaping direction, if exists}
    \vspace{2mm}
    If any of the following candidates lies in the feasible set and decreases the objective by $\Omega(\delta)$, return it: \\
    \textbf{Case 1. Large gradient}: $\argmin_{\vy \in \s{S} \cap \ball(\s{r})} \vy^\top \s{\vv}$ \\
    \textbf{Case 2. Solution in the interior}: $\vy \gets -\s{\vM}^{-1} \s{\vv}$ \\
    \textbf{Case 3. Solution on the boundary}: \\
    \begin{algbox}
    Find orthogonal $\vQ$ and diagonal $\vLambda = \diag(\lambda_1, \ldots, \lambda_{\dim \act})$ such that $\s{\vM} \approx \vQ^\top \vLambda \vQ$ \\
    Let $\tilde \vv \gets \vQ \vv$ \\
    We consider points with coordinates $y_i \gets \frac{\tilde v_i}{\mu_j - \lambda_i}$ for some $\mu$ \\
    Find the values of $\mu$ for which the points have norm $\s{r}$ \\
    The candidates are the points corresponding to these values of $\mu$
    \end{algbox}
    If no candidate satisfies the condition, return $\bot$
\end{algorithm}

Our main result is the following theorem that shows how to find a $\delta$-SOSP.

\begin{theorem}
    \label{thm:general_case}
    Let $S = \{\vx \mid \vA \vx \le \vb\}$ be a set defined by an intersection of $k$ linear inequality constraints.
    Let $f$ satisfy Assumptions~\ref{ass:lipschitz} and~\ref{ass:sgd} and let $\min_{\vx \in S} f(\vx) = f^\star$.
    Then there exists an algorithm which for $\delta > 0$ finds a $\delta$-SOSP in $\tilde O(\frac{f(\vx_0) - f^\star}{\delta}d^3 ( 2^{k} + \frac{\sigma^2}{\delta^{\nicefrac{4}{3}}}))$ time using $\tilde O(\frac{f(\vx_0) - f^\star}{\delta}(d + \frac{d^3 \sigma^2}{\delta^{\nicefrac{4}{3}}}))$ stochastic gradient oracle calls.
\end{theorem}

Our approach is outlined in Algorithms~\ref{alg:escaping_general} and~\ref{alg:eigenvector_general}\footnote{Algorithm~\ref{alg:escaping_general} and~\ref{alg:eigenvector_general} are simplified versions of rigorous Algorithms~\ref{alg:escaping_general_full} and~\ref{alg:eigenvector_general_full} from Appendix~\ref{app:general}}:
\begin{itemize}
    \item If $\vx$ is not a $\delta$-SOSP, Algorithm~\ref{alg:escaping_general} finds an escape direction,
        i.e. a point $\vy \in S \cap \ball(\vx, r)$ which significantly decreases the objective value: $f(\vy) < f(\vx) - \Omega(\delta)$.
        Therefore, if $\vx_0$ is the initial point, our algorithm requires $O(\frac{f(\vx_0) - f^\star}{\delta})$ calls of Algorithm~\ref{alg:escaping_general}.
    \item 
        Algorithm~\ref{alg:escaping_general} enumerates all possible sets of active constraints
\end{itemize}
Recall that we aim to find a $\delta$-SOSP, i.e. $\vx \in S$ such that $f(\vy) \ge f(\vx) - \delta, \forall \vy \in S \cap \ball(\vx, r)$, where $r=\trueR$.

Similarly to Section~\ref{sec:copositivity}, we use a quadratic approximation and consider two cases depending on whether the minimizer $\vy$ lies in the interior or on the boundary of $S$.
The second case can be reduced to the first in a way similar to Section~\ref{sec:copositivity}.
To find the minimizer in the interior, we consider the following cases in Algorithm~\ref{alg:escaping_general}:
    
\textbf{Case 1.}    When the gradient is large, we ignore the quadratic term, and optimize the function based on the gradient.
        This covers the situation when the objective is sensitive to the change in the argument.
    
\textbf{Case 2.} When the optimum lies in $\interior{\ball(\vx, r)}$, none of the constraints are active,
        and hence we can find the unique unconstrained critical point directly.

\textbf{Case 3.} When the optimum lies in $\boundary{\ball(\vx, r)}$, the only active constraint is $c(\vy) = \|\vy - \vx\|^2 - r^2 = 0$.
        By KKT conditions, for $\vy$ to be the minimizer, there must exist $\mu$ such that $\nabla f(\vy) = \mu \nabla c(\vy)$.
        We show that for each $\mu$, there exists a unique $\vy(\mu)$ satisfying the condition above,
        and only for $O(d)$ values of $\mu$ we have $\vy(\mu) \in \boundary{\ball(\vx, r)}$, resulting in $O(d)$ candidate solutions.

We now outline the proof of Theorem~\ref{thm:general_case} (full proof is in Appendix~\ref{app:general}).
As we show in Section~\ref{ssec:quad_approx}, we can consider a quadratic approximation of the objective.
By guessing which constraints are active at the minimizer $\vx^\star$ and enforcing these constraints, we restrict the function to some affine subspace $\act$.
By parameterizing $\act$, we eliminate enforced constraints, and, since the rest of the constraints are not active at $\vx^\star$, we need to optimize a quadratic function in the intersection of a ball and linear inequality constraints (see Section~\ref{ssec:projection}).


\subsection{Algorithm~\ref{alg:escaping_general}: Quadratic approximation}
\label{ssec:quad_approx}

The goal is to build a quadratic approximation of the objective with some additional properties (see below).
To simplify the presentation, w.l.o.g. (by shifting the coordinate system) we assume that the current saddle point is $\vzero$ and consider the quadratic approximation of the function:
\[
    f(\vx) \approx f(\vzero) + \vx^\top \nabla f(\vzero) + \frac 12 \vx^\top \nabla^2 f(\vzero) \vx
.\]
Since $f$ is $\rho$-Lipschitz and $r=\trueR$, in $\ball(r)$ the quadratic approximation deviates from $f$ by at most $\frac{\delta}{6}$ (see derivation before Definition~\ref{def:sosp}).
For a small value $\xi$ (to be specified later), we instead analyze a noisy function
\[
    f'(\vx) =  f(\vzero) + \vx^\top \vv + \frac 12 \vx^\top \vM \vx
,\]
where:
\begin{enumerate}
    \item
        $\vv$ is a perturbed approximation of $\nabla f(\vzero)$.
        The perturbation guarantees that w.h.p. all coordinates of $\vv$ are sufficiently separated from $0$ and linear systems of the form $(\vM - \mu I) \vx = \vv$ with $\rank (\vM - \mu I) < d$ don't have any solutions, simplifying the analysis.
        To approximate the gradient, we use algorithm $\vrsg$ (see Appendix~\ref{app:general}, Lemma~\ref{lem:vrsg}), which w.h.p. estimates the gradient with precision $\tilde \sigma$ using $\tilde O(\frac{\sigma^2}{\tilde \sigma^2})$ stochastic gradient oracle calls.
    \item
        $\vM$ is a perturbed approximation of $\nabla^2 f(\vzero)$.
        The perturbation guarantees that $\vM$ is non-degenerate with probability $1$.
        Since the $i$-th column of $\nabla^2 f(\vzero)$ is by definition $\lim\limits_{\tau \to 0} \frac{\nabla f(\tau \ve_i) - \nabla f(\vzero)}{\tau}$, using a sufficiently small $\tau$ and approximating $\nabla f(\tau \ve_i)$ and $\nabla f(\vzero)$ using $\vrsg$, we find good approximation of the Hessian.
\end{enumerate}
Combining this with the derivation before Definition~\ref{def:sosp}, we show the following:
\begin{lemma}
    \label{lem:stochastic-approximation}
    Let $f$ satisfy Assumptions~\ref{ass:lipschitz} and~\ref{ass:sgd}.
    Let $f'(\vx) = f(\vzero) + \vx^\top \vv + \frac 12 \vx^\top \vM \vx$, where $\vv$ and $\vM$ are as in Algorithm~\ref{alg:escaping_general}.
    For $\delta > 0$, $r = \trueR$ we have $\|f'(\vx) - f(\vx)\| < \frac{\delta}{2}$ for all $\vx \in \ball(r)$ w.h.p.
\end{lemma}

\paragraph{Reducing Case \texorpdfstring{$\vx^\star \in \boundary{S}$}{v* on the boundary of S} to Case \texorpdfstring{$\vx^\star \in \interior{S}$}{v* in Int S}.}

Similarly to Section~\ref{sec:copositivity}, we reduce the case $\vx^\star \in \boundary{S}$ to the case $\vx^\star \in \interior{S}$.
If $\vx^\star \in \boundary{S}$, then there exist a non-empty set $\actInd$ of constraints active at $\vx^\star$.
Consider the iteration of Algorithm~\ref{alg:escaping_general} where constraints from $\actInd$ are active.
These active constraints define an affine subspace $\act$, which e parameterize:
    if $\vp = \proj_\act (\vx)$ and $\vO \in \R^{d \times \dim \act}$ is an orthonormal basis of $\act$,
    then any point in $\act$ can be represented as $\vp + \vO \vy$ for $\vy \in \R^{\dim \act}$.
Defining $g(\vy) = f'(\vp + \vO \vy)$,
minimizing $f'$ in $\act \cap S \cap \ball(r)$ is equivalent to minimizing $g$ in $\s{S} \cap \ball(\s{r})$, where:
\begin{enumerate}
    \item $\s{S}$ is a set of points $\vy \in \R^{\dim \act}$ such that $\vp + \vO \vy \in S$, namely $\s{S} = \{\vy \mid \vA_i (\vp + \vO \vy) \le b_i,\ i \notin \actInd\}$.
        Hence, $\s{S}$ is defined by linear inequalities, similarly to $S$.
    \item $\s{r}$ is a radius such that condition $\vy \in \ball(\s{r})$ is equivalent to $\vp + \vO \vy \in \ball(r)$.
        Since $\vp$ is the projection of $\vzero$ on $\act$, we have $\vO^\top \vp = \vzero$, and hence $\|\vp + \vO \vy\|^2 = \|\vp\|^2 + \|\vO \vy\|^2$.
        Since $\vO$ is an orthonormal basis of $\act$, $\|\vO \vy\| = \|\vy\|$, and hence $\s{r} = \sqrt{r^2 - \|\vp\|^2}$.
\end{enumerate}
For $\vy^\star$ such that $\vx^\star = \vp + \vO \vy^\star$, no constraints from $\s{S}$ are active, and hence $\vx^\star \in \interior{\s{S}}$.

\subsection{Algorithm~\ref{alg:eigenvector_general}: Escaping when \texorpdfstring{$\vy^\star \in \interior{\s{S}}$}{y* in Int S}}
\label{ssec:interior_cases}

In this section, we the find minimizer $\vy^\star$ of function $g(\vy) = \frac 12 \vy^\top \s{\vM} \vy + \vy^\top \s{\vv} + C$ in $\s{S} \cap \ball(\s{r})$, while assuming that $\vy^\star \in \interior{\s{S}}$.
Since the solutions we find can be approximate, we have to guarantee that the objective is not too sensitive to the change of its argument.
It suffices to consider the case when $\|\vv_\bot\|$ is bounded, since for any $\vy \in \ball(\s{r})$ and perturbation $\vh$ there exists $\tau \in [0, 1]$ such that:
\begin{align*}
    |g(\vy) - g(\vy + \vh)|
    &= |(\nabla g(\vy + \tau \vh))^\top \vh| \\
    &\le (\|\nabla g(\vzero)\| + L \|\vy + \tau \vh\|) \|\s{\vh}\|\\
    &\le (\|\vv_\bot\| + L (\s{r} + \|\vh\|)) \|\vh\|
,\end{align*}
where we used that the objective is $L$-smooth and hence $\|\nabla g(\vy + \tau \vh)\| \le \|\nabla g(\vzero)\| + L \|\vy + \tau \vh\|$.
We consider the situation when $\|\s{\vv}\|$ is large as a separate case.
Otherwise, for $\vy^\star$, there are only two options: either $\vy^\star \in \interior{\ball(\s{r})}$ or $\vy^\star \in \boundary{\ball(\s{r})}$.
Algorithm~\ref{alg:eigenvector_general} handles these cases, as well as the case when $\|\vv\|$ is large, separately. 
\begin{figure}[ht!]
    \centering
    \begin{subfigure}[t]{\figwidth}
        \centering
        \begin{tikzpicture}[scale=\figscale]
        \draw[blue] (1,0) -- ++(180:4) -- +(90:4);
        \filldraw[fill=blue!3, draw=blue] (0,0) arc (0:90:3) -- (-3,0) -- cycle;
        \draw[red, line width=\figlinethick] (1,0) -- ++(180:4) -- +(90:4);
        \end{tikzpicture}
        \caption{Case 0: the minimizer is on $\boundary{S}$. We handle this case when processing the corresponding set of active constraints in Algorithm~\ref{alg:escaping_general}.}
    \end{subfigure}
    \quad
    \begin{subfigure}[t]{\figwidth}
        \centering
        \begin{tikzpicture}[scale=\figscale]
        \draw[blue] (1,0) -- ++(180:4) -- +(90:4);
        \filldraw[fill=blue!3, draw=blue] (0,0) arc (0:90:3) -- (-3,0) -- cycle;
        \draw[red, -stealth] (-3,0) -- ++(60:5);
        \end{tikzpicture}
        \caption{Case 1: large linear term. Then, we optimize the objective based on the linear term alone. This case is separate to avoid numerical issues in other cases}
    \end{subfigure}
    \quad
    \begin{subfigure}[t]{\figwidth}
        \centering
        \begin{tikzpicture}[scale=\figscale]
        \draw[blue] (1,0) -- ++(180:4) -- +(90:4);
        \filldraw[fill=red!10, draw=blue] (0,0) arc (0:90:3) -- (-3,0) -- cycle;
        \end{tikzpicture}
        \caption{Case 2: the minimizer is in the interior. Then, we can find the critical point analytically, by solving a linear equation.}
    \end{subfigure}
    \quad
    \begin{subfigure}[t]{\figwidth}
        \centering
        \begin{tikzpicture}[scale=\figscale]
        \draw[blue] (1,0) -- ++(180:4) -- +(90:4);
        \filldraw[fill=blue!3, draw=blue] (0,0) arc (0:90:3) -- (-3,0) -- cycle;
        \draw[red, line width=\figlinethick] (0,0) arc (0:90:3);
        \end{tikzpicture}
        \caption{Case 3: the minimizer is on the boundary. Then, there are at most $O(d)$ candidates with norm $\s{r}$}
    \end{subfigure}
    \quad

    \caption{Cases of Algorithm~\ref{alg:eigenvector_general}} \label{fig:escape-cases}
\end{figure}
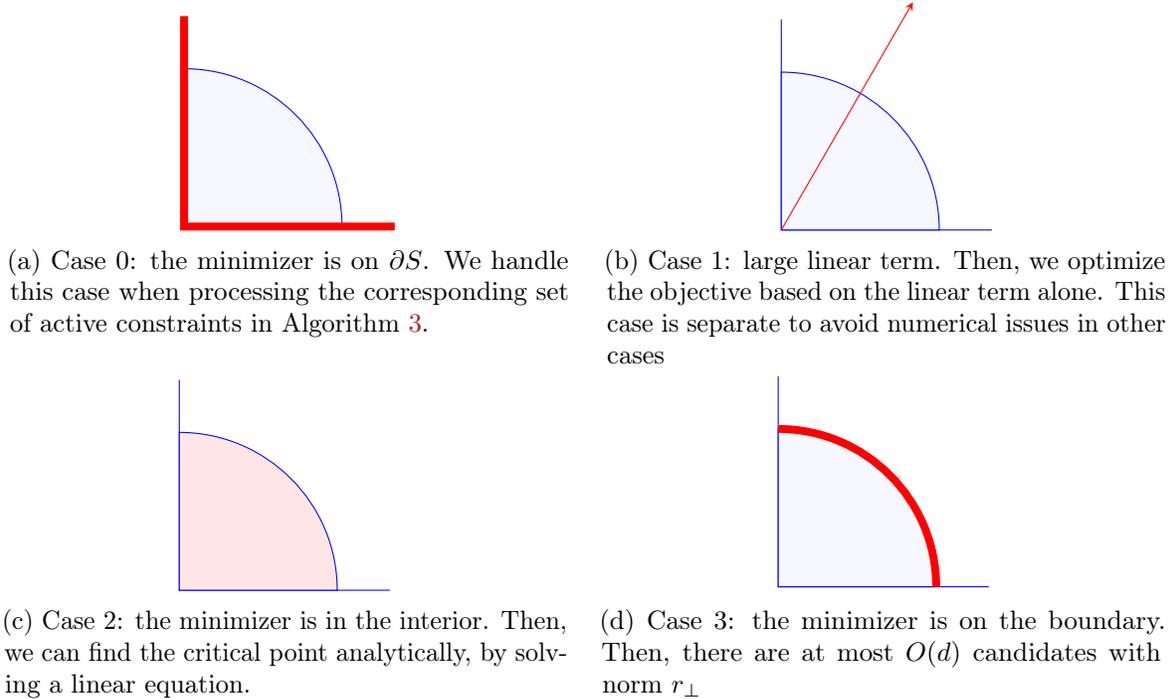

\paragraph{Case 1: $\|\s{\vv}\|$ is large.}
If $\|\s{\vv}\|$ is large and we can find $\vy$ with small $\vy^\top \s{\vv}$, the linear term alone suffices to improve the objective.
We show that, if such $\vy$ doesn't exist, then $g(\vy^\star)$ requires $\vy^\star \in \boundary{\s{S}}$, which contradicts that $\vy^\star \in \interior{\s{S}}$.
Below we assume that $\|\s{\vv}\|$ is bounded.

\paragraph{Case 2: $\vy \in \interior{\ball(\s{r})}$.}
In this case, $\vy^\star$ is an unconstrained critical point of $g$, and hence it must satisfy $\nabla g(\vy) = \vzero$, implying $\s{\vM} \vy + \s{\vv} = \vzero$ which gives the unique solution $\vy = - \s{\vM}^{-1} \s{\vv}$.
since $\vM$ is a perturbed matrix, so is $\s{\vM}$, and hence $\s{\vM}$ is non-degenerate with probability $1$.
It remains to verify that $\vy \in \ball(\s{r}) \cap \s{S}$ and $\vy$ decreases the objective by $\Omega(\delta)$.

\paragraph{Case 3: $\vy \in \boundary{\ball(\s{r})}$.}
Since the only active constraint at $\vy^\star$ is $c(\vy) = \frac 12(\|\vy\|^2  - \s{r}^2) = 0$,
    by the KKT conditions, any critical point must satisfy $\nabla g(\vy) = \mu \nabla c(\vy)$ for some $\mu \in \R$,
    which is equivalent to $\s{\vM} \vy + \s{\vv} = \mu \vy$.
    Hence, for any fixed $\mu$, $\vy$ must be a solution of linear system $(\s{\vM} - \mu I) \vy = -\s{\vv}$.
    When $\s{\vM} - \mu I$ is degenerate (i.e. $\mu$ is an eigenvalue of $\s{\vM}$), due to perturbation of $\s{\vv}$, the system doesn't have any solution with probability $1$.
    It leaves us with the case when $\s{\vM} - \mu I$ is non-degenerate, when there exists a unique solution $\vy(\mu) := - (\s{\vM} - \mu I)^{-1} \s{\vv}$.
    
    \textbf{Diagonalization.} Dependence of $(\s{\vM} - \mu I)^{-1}$ on $\mu$ is non-trivial, but for a diagonal $\s{\vM}$, the inverse can be found explicitly.
    Hence, we perform diagonalization of $\s{\vM}$: we find orthogonal $\vQ$ and diagonal $\vLambda$
        such that $\|\s{\vM} - \vQ^\top \vLambda \vQ\| < \eps$ in time $O(d^3 \log \nicefrac 1\eps)$~\citep{PC99}.
    Setting $\eps = O(\nicefrac{\delta}{\s{r}^2})$, we guarantee that the function changes by at most $O(\delta)$ in $\ball(\s{r})$.
    The function $\vy(\mu) = - (\vQ^\top \vLambda \vQ - \mu I)^{-1} \s{\vv}$ can be written as $\vQ \vy(\mu) = -(\vLambda - \mu I) \vQ \s{\vv}$, and hence we work with rotated vectors $\tilde \vy(\mu) := \vQ \vy(\mu)$ and $\tilde \vv := \vQ \s{\vv}$.
    
    \textbf{Finding candidate $\mu$.} Since $\tilde \vy(\mu) = - (\s{\vM} - \mu I)^{-1} \tilde \vv$, for the $i$-th coordinate of $\tilde \vy(\mu)$ we have
        $\tilde y_i(\mu) = \frac{\tilde v_i}{\mu - \lambda_i}$.
    Since we are only interested in $\vy(\mu) \in \boundary{\ball(\s{r})}$ and $\vQ$ is an orthogonal matrix, we must have:
    \[
        \|\s{r}\|^2
        = \|\vy(\mu)\|^2
        = \|\tilde \vy(\mu)\|^2
        = \sum_{i=1}^d \tilde y_i(\mu)^2
        = \sum_i \frac{\tilde v_i^2}{(\mu - \lambda_i)^2}
    \]
    After multiplying the equation by $\prod_i (\mu - \lambda_i)^2$, we get an equation of the form $p(\mu) = 0$,
        where $p$ is a polynomial of degree $2d$.
    We find roots $\mu_1, \ldots, \mu_{2d}$ of the polynomial in time $O(d^2 \log d \cdot \log \log \nicefrac 1\eps)$~\citep{PAN1987591},
        where $\eps$ is the required root precision.
    For each $i$, we compute $\vy(\mu_i)$ and verify whether it lies in $\ball(\s{r}) \cap \s{S}$ and improves the objective by $-\Omega(\delta)$.
        
    \textbf{Precision.} Since the roots of the polynomial are approximate, when $\mu$ is close to $\lambda_i$,
        even a small perturbation of $\mu$ can strongly affect $y_i(\mu) = \frac{\tilde v_i}{\mu - \lambda_i}$.
    We solve this as follows: since $\|\tilde \vy(\mu)\| = \s{r}$,
        for each $i$ we must have $|\tilde y_i(\mu)|\le \s{r}$, implying $|\mu - \lambda_i| \ge \frac{|\tilde v_i|}{\s{r}}$.
    Therefore, $\mu$ must be sufficiently far from any $\lambda_i$, where the lower bound on the distance depends on $\s{r} \le \trueR$ and on $|\tilde v_i|$.
    By adding noise to $\vv$ ,
        we guarantee that the noise is preserved in $\tilde \vv$ so that each coordinate is sufficiently separated from $0$ w.h.p. This is formalized in Appendix~\ref{app:general}, Lemma~\ref{lem:solve_general}.
\section{Conclusion}

In this paper, we have shown that it’s possible to escape from a constrained second-order stationary point with the logarithmic number of constraints within polynomial time and using only a polynomial number of stochastic gradient oracle calls.
We provide experimental results in Appendix~\ref{app:experiments}.

An open question is to determine the conditions that on one hand guarantee escaping from a saddle point in polynomial time even for the linear number of constraints, and on the other hand hold in practice. One such condition can be strict complementarity.

Another open question is handling non-linear constraints. We believe that it can be straightforwardly achieved using techniques from~\citet{GHJY15} by using assumptions on curvature and linear independence of the constraints. Finally, an interesting question would be a simpler algorithm for the general case, e.g. an algorithm resembling the approach from Section~\ref{sec:copositivity}.

\ifarxiv
    \bibliographystyle{plainnat}
\else
    \bibliographystyle{named}
\fi
\bibliography{main}

\newpage
\appendix

\section{Standard Facts}
\label{app:facts}

\begin{fact}[Efficient projection on a polyhedron]
    For an arbitrary point $\vx \in \R^d$ and a set $S = \{\vx \in \R^d \mid \vA \vx \le \vb\}$~--
    an intersection of $k$ linear inequality constraints, we can find a projection of $\vx$ onto $S$ in time $\poly(d)$.
\end{fact}

\begin{fact}[Convergence of power iteration]
    \label{fact:power_convergence}
    Let $\vM \in \R^{d \times d}$ be a PSD matrix with largest eigenvalue $\lmax$.
    Let a sequence $\{\vx_t\}_{t=0}^\infty$ in $\R^d$ be defined according to the power iteration: $\vx_0 \sim \mathcal N(\vzero, I)$ and $\vx_{t+1} = \vM \vx_t$.
    Then for $\delta < \lmax$ we have $\frac{\vx_T^\top \vM \vx_T}{\|\vx_T\|^2} > \delta$ with probability $1 - O(\xi)$
        after $T=O \left(\frac{\log (d \delta) - \log (\xi (\lmax - \delta))}{\log \nicefrac{\lmax}{\delta}}\right)$ iterations.
\end{fact}
\begin{proof}
    If $\lambda_i < \delta$, the corresponding component of the vector decreases exponentially with rate at least $\nicefrac{\delta}{\lmax}$ and are eventually dominated by the component corresponding to $\lmax$.
    This leaves us only with components corresponding to $\lambda_i \ge \delta$, finishing the proof.
    
    Let $\lambda_1, \ldots, \lambda_d$ be the eigenvalues of $\vM$ and let $\ve_1, \ldots, \ve_d$ be the corresponding eigenvectors.
    Then $\vx_0 = \sum_{i=1}^d \InnerProd{\vx_0, \ve_i} \ve_i$ and $\vx_T = \sum_{i=1}^d \InnerProd{\vx_0, \ve_i} \lambda_i^T \ve_i$.
    Hence,
    \begin{align*}
        \frac{\vx_T^\top \vM \vx_T}{\|\vx_T\|^2}
        &= \frac{\sum_{i=1}^d \InnerProd{\vx_0, \ve_i}^2 \lambda_i^{2T}\lambda_i}{\sum_{i=1}^d \InnerProd{\vx_0, \ve_i}^2 \lambda_i^{2T}} \\
        &= \frac{\sum_{i=1}^d \InnerProd{\vx_0, \ve_i}^2 (\nicefrac{\lambda_i}{\lmax})^{2T} \lambda_i} 
            {\sum_{i=1}^d \InnerProd{\vx_0, \ve_i}^2 (\nicefrac{\lambda_i}{\lmax})^{2T}} \\
        &= \delta + \frac{\sum_{i=1}^d \InnerProd{\vx_0, \ve_i}^2 (\nicefrac{\lambda_i}{\lmax})^{2T} (\lambda_i - \delta)} 
            {\sum_{i=1}^d \InnerProd{\vx_0, \ve_i}^2 (\nicefrac{\lambda_i}{\lmax})^{2T}}
    .\end{align*}
    It suffices to show that the numerator in the last term is non-negative.
    Clearly, in the summation, the terms with $\lambda_i \ge \delta$ are positive and the terms with $\lambda_i < \delta$ are negative.
    Hence, we must lower-bound $\InnerProd{\vx_0, \ve_i}^2$ when $\lambda_i \ge \delta$ and upper-bound $\InnerProd{\vx_0, \ve_i}^2$ when $\lambda_i < \delta$.
    Since $\vx_0 \sim \mathcal N(\vzero, I)$, the following holds for all $i \in [d]$:
    \begin{itemize}
        \item With probability $1 - O(\nicefrac{\xi}{d})$, $|\InnerProd{\vx_0, \ve_i}| \ge \nicefrac{\xi}{d}$.
        \item With probability $1 - O(\nicefrac{\xi}{d})$, $|\InnerProd{\vx_0, \ve_i}| \le \log \nicefrac{d}{\xi}$.
    \end{itemize}
    By the union bound, these inequalities hold for all $i$ with probability $1 - O(\xi)$.
    For $i$ such that $\lambda_i=\lmax$, we have
    \[
        \InnerProd{\vx_0, \ve_d}^2 (\nicefrac{\lambda_i}{\lmax})^{2T} (\lmax - \delta) 
        \ge (\nicefrac{\xi}{d})^2 (\lmax - \delta) 
    .\]
    For all $i$ such that $\lambda_i < \delta$, we have:
    \[
        \InnerProd{\vx_0, \ve_i}^2 (\nicefrac{\lambda_i}{\lmax})^{2T} (\lambda_i - \delta) 
        \ge - \delta (\nicefrac{\delta}{\lmax})^{2T} \log^2 \nicefrac{d}{\xi}
    .\]
    Since there are at most $d$ such terms, to guarantee that $\sum_{i=1}^d \InnerProd{\vx_0, \ve_i}^2 (\nicefrac{\lambda_i}{\lmax})^{2T} (\lambda_i - \delta) \ge 0$, it suffices to have:
    \[
        (\nicefrac{\xi}{d})^2 (\lmax - \delta) \ge d \delta (\nicefrac{\delta}{\lmax})^{2T} \log^2 \nicefrac{d}{\xi}
    ,\]
    which holds when
    \begin{align*}        
        T
        &\ge \frac{\log \left(\frac{d \delta}{(\lmax - \delta)} (\nicefrac{d}{\xi})^2 \log^2 \nicefrac{d}{\xi}\right)}{2 \log \nicefrac{\lmax}{\delta}} \\
        &= O \left(\frac{\log (d \delta) - \log (\xi (\lmax - \delta))}{\log \nicefrac{\lmax}{\delta}}\right)
    .
    \end{align*}
\end{proof}

\begin{fact}[Coordinate-wise median trick]
    \label{fact:median_trick}
    Let $\xi^{(1)}, \ldots, \xi^{(n)}$ be independent random variables in $\R^d$ sampled from the same distribution with mean $\E[\xi^{(i)}] = \mu$ and variance $\E [\|\xi^{(1)} - \mu\|^2] \le \sigma^2$.
    If $\zeta$ is a coordinate-wise median of $\xi_1, \ldots, \xi_n$, then
    \[
        \Pr [\|\zeta - \mu\| > 2 \sigma] = d \cdot e^{-\Omega(n)}
    .\]
\end{fact}
\begin{proof}
    The proof idea is to apply median trick to each coordinate separately.
    For each $\xi^{(i)}$ we have:
    \[
        \E [\|\xi^{(i)} - \mu\|^2]
        = \E [\sum_{j=1}^d (\xi^{(i)}_j - \mu_j)^2 ]
        = \sum_{j=1}^d \E [(\xi^{(i)}_j - \mu_j)^2]
    .\]
    For each coordinate $j$, let $\sigma_j^2 = \E [(\xi^{(i)}_j - \mu_j)^2]$.
    Using Chebyshev's inequality, we have:
    \[
        \Pr [(\xi^{(i)}_j - \mu_j)^2 \ge 4 \sigma_j^2] \le \frac 14
    .\]
    Let $X_{i,j}$ be random variables such that $X_{i,j} = 1$ if $(\xi^{(i)}_j - \mu_j)^2 \ge 4 \sigma_j^2$ and $X_{i,j} = 0$ otherwise.
    From the above inequality, we have $\E[\sum_{i=1}^n X_{i,j}] \le \frac{n}{4}$.
    Since $X_{1,j}, \ldots, X_{n,j}$ are bounded and independent, by the Chernoff bound we have:
    \[
        \Pr [\sum_{i=1}^n X_{i,j} > \frac{n}{2}] = e^{-\Omega(n)}
    .\]
    Hence, for any $j$, with probability $1 - e^{-\Omega(n)}$ at most half of $X_{i,j}$ are $1$.
    Therefore, at least half of $\xi^{(1)}_j, \ldots, \xi^{(n)}_j$ lie in $[\mu_j - \sigma_j, \mu_j + \sigma_j]$, and hence the same holds for their median $\zeta_j$.
    Taking the union bound over all coordinates, with probability $1 - d \cdot e^{-\Omega(n)}$ we have:
    \[
        \|\zeta - \mu\|^2
        = \sum_{j=1}^d (\zeta_j - \mu_j)^2
        \le \sum_{j=1}^d 4 \sigma_j^2
        \le 4 \sigma^2
    .\]
\end{proof}

\begin{fact}
    For a fixed $\delta$, let $\eps = \sqrt[3]{\delta^2 \rho}$. If $\|\nabla f(\vx)\| < \eps$ and $\lmin(\nabla^2 f(\vx)) > -\sqrt{\rho \eps}$, for any $\vh \in \ball(\vx, r)$ with $r=\sqrt[3]{\nicefrac{\delta}{\rho}}$ we have:
    \[
        f(\vx + \vh) - f(\vx)
        \ge \vh^\top \nabla f(\vx) + \frac 12 \vh^\top \nabla^2 f(\vx) \vh - \frac{\rho}{6} \|\vh\|^3
        \ge - 2 \delta
    \]
\end{fact}
\begin{proof}
    By the Taylor expansion, we know:
    \[
        f(\vx + \vh) - f(\vx)
        \ge \vh^\top \nabla f(\vx) + \frac 12 \vh^\top \nabla^2 f(\vx) \vh - \frac{\rho}{6} \|\vh\|^3
    .\]
    It remains to bound each term separately.
    Note that for this choice of $\eps$, we have $\sqrt{\rho \eps} = \sqrt[3]{\delta \rho^2}$.
    Hence:
    \begin{align*}
        |\vh^\top \nabla f(\vx)| &\le r \eps
            &&\le \sqrt[3]{\nicefrac{\delta}{\rho}} \cdot \sqrt[3]{\delta^2 \rho} &= \delta \\
        \frac 12 \vh^\top \nabla^2 f(\vx) \vh &\ge - \frac 12 r^2 \sqrt{\rho \eps}
            &&\ge \frac 12 - \sqrt[3]{\nicefrac{\delta^2}{\rho^2}} \cdot \sqrt[3]{\delta \rho^2} &= -\frac{\delta}{2} \\
        \frac{\rho}{6} \|\vh\|^3 &\le \frac 16 \rho r^3
            &&\le \frac 16 \rho \cdot \nicefrac{\delta}{\rho} &= \frac{\delta}{6}
    .\end{align*}
    Combining these inequalities gives the required bound.
\end{proof}
\section{Missing Proofs from Section~\ref{sec:copositivity}}
\label{app:copositivity}

\begin{definition}
    A set $S$ is a \emph{linear cone} if $\vx \in S$ implies $\alpha \vx \in S$ for all $\alpha \ge 0$.
\end{definition}
If $S$ is a linear cone, then scaling a vector preserves its belonging to $S$.
In particular, $S = \{\vx \mid \vA \vx \le 0\}$ is a linear cone.

\begin{lemma}[Lemma~\ref{lemma:copositivity_check}]
    Let $f(\vx) = \frac 12 \vx^\top \vM \vx$ and $S$ be a closed linear cone.
    If $\vzero$ is not a $(\delta,r)$-boundary QCSP for $\delta, r > 0$, then the following statements are equivalent:
    \begin{compactenum}
        \item $\vzero$ is $(\delta,r)$-QCSP, i.e. $\min_{\vx \in S \cap \ball(r)} f(\vx) < -\delta$.
        \item There exists an eigenvector $\ve$ of $\vM$ such that $\ve \in \interior{S} \cap \boundary{\ball(r)}$ and $f(\ve) < -\delta$.
    \end{compactenum}
\end{lemma}

\begin{proof}
    $ $ \\
    \noindent $2) \implies 1)$ follows trivially by definition of $(\delta,r)$-QCSP.
    
    \noindent $1) \implies 2)$.
    We show that minimizer $\vx^\star$ lies in $\boundary(\ball(r))$, and, since it's a local minimum, by the method of Lagrangian multipliers, we show $\nabla f(\vx^\star) = \lambda \vx^\star$ for some $\lambda$, where $\nabla f(\vx^\star) = \vM \vx^\star$.
    
    Let $\vx^\star$ be the minimizer of $f$ on $S \cap \ball(r)$.
    Then $\vx^\star \in \boundary{\ball(r)}$, since otherwise we can rescale the minimizer to be on the boundary.
    I.e. for $\vy = \frac{r}{\|\vx^\star\|} \vx^\star$ we have $f(\vy) = \frac{r^2}{\|\vx^\star\|^2} f(\vx^\star) < f(\vx^\star)$. Furthermore, $\vy \in S$ since $\vx^\star \in S$ and $S$ is a linear cone, and $\vy \in \boundary{\ball(r)}$ since $\left\|\vy\right\| = r$.
    
    From the above reasoning we have that $\vx^\star \in S \cap \boundary{\ball(r)}$,
        and by the assumption of the lemma we have $f(\vx^\star) < - \delta$.
    Since
    \[\min_{\vx \in \boundary{S} \cap \boundary{\ball(r)}} f(\vx) \ge \min_{\vx \in \boundary{S} \cap \ball(r)} f(\vx) \ge -\delta,\]
    the minimizer $\vx^\star$ lies in $\interior{S} \cap \boundary{\ball(r)}$.
    Since $\vx^\star \in \boundary{\ball(r)}$, it satisfies the constraint $g(\vx^\star) = \|\vx^\star\|^2 - r^2 = 0$,
        and by the method of Lagrangian multipliers we must have $\nabla f(\vx^\star) = \lambda \nabla g(\vx^\star)$ or equivalently $\vM \vx^\star = \lambda \vx^\star$ for some $\lambda$.
    Therefore, $\vx^\star$ is an eigenvector of $\vM$ and $f(\vx^\star) < -\delta$.

\end{proof}

\begin{lemma}[Lemma~\ref{lemma:copositivity_approx_eigen}]
    Let $f(\vx) = \frac 12 \vx^\top \vM \vx$ and $S$ be a closed linear cone.
    For $\delta, r > 0$ and $\hat{\vx} \in \boundary{\ball(r)}$, if the following conditions hold, then either $\hat{\vx} \in S$ or $-\hat{\vx} \in S$:
    \begin{compactenum}
        \item $\vzero$ is a $(\delta,r)$-QCSP, i.e. $\min_{\vx \in S \cap \ball(r)} f(\vx) < -\delta$,
        \item $\vzero$ is not a $(\delta,r)$-boundary QCSP, i.e. $\min_{\vx \in \boundary{S} \cap \ball(r)} f(\vx) \ge -\delta$,
        \item $f(\hat{\vx}) < - \delta$.
    \end{compactenum}
\end{lemma}

\begin{proof}
    We know that there exists an eigenvector $\ve \in S$ with $f(\ve) < -\delta$.
    If both $\hat{\vx}$ and $-\hat{\vx}$ don't belong to $S$, we consider an arc on $\boundary{\ball(r)}$ connecting $\ve$ and $\hat{\vx}$ (or $-\hat{\vx}$, see below).
    We will show that for any point $\vx$ on the arc, $f(\vx) < -\delta$, and, since the arc intersects with $\boundary{S}$, this contradicts assumption that $\vzero$ is not a $(\delta,r)$-boundary QCSP.
    
    For contradiction, assume that both $-\hat{\vx}$ and $\hat{\vx}$ don't belong to $S$.
    From Lemma~\ref{lemma:copositivity_check} we know that there exists an eigenvector $\ve$ with $f(\ve) < -\delta$ and $\ve \in S \cap \boundary{\ball(r)}$.
    Either $\hat{\vx}^\top \ve \ge 0$ or $-\hat{\vx}^\top \ve \ge 0$, and w.l.o.g. we assume $\hat{\vx}^\top \ve \ge 0$.
    Consider an arc between $\hat{\vx}$ and $\ve$ on $\boundary{\ball(r)}$:
    \[
        U=\{\alpha \hat{\vx} + \beta \ve \mid \alpha,\beta \ge 0\} \cap \partial{\ball(r)}
    .\]
    Since $U$ is a connected set, $\hat{\vx} \notin S$ and $\ve \in S$, there exists $\vx \in U \cap \boundary{S}$.
    Since $\vx = \alpha \hat{\vx} + \beta \ve$ for some $\alpha,\beta \ge 0$, we have:
    \begin{align*}
        \frac 12 \vx^\top \vM \vx
        &= \frac 12 (\alpha \hat{\vx} + \beta \ve)^{\top} \vM (\alpha \hat{\vx} + \beta \ve) \\
        &= \frac 12 \alpha^2 \hat{\vx}^{\top} \vM \hat{\vx} + \alpha \beta \ve^\top \vM \hat{\vx} + \frac 12\beta^2 \ve^\top \vM \ve
    .\end{align*}
    By our assumption, $\frac 12 \hat{\vx}^\top \vM \hat{\vx} < -\delta$ and $\frac 12 \ve^\top \vM \ve < -\delta$.
    Since $\hat{\vx}^\top \ve \ge 0$ and $\ve$ is an eigenvector of $\ve$ with eigenvalue at most $\frac{f(\ve)}{\nicefrac{\|\ve\|^2}{2}}=-\frac{2\delta}{r^2}$, we have
    \[
        \alpha \beta \ve^\top \vM \hat{\vx} < -\frac{2\delta}{r^2} \alpha \beta \ve^\top \hat{\vx}
    ,\]
    where we used that all terms are non-negative.
    Finally, using $\|\hat{\vx}\|=\|\ve\|=\|\alpha \hat{\vx} + \beta \ve\|=r$:
    \begin{align*}
        \frac 12 \vx^\top \vM \vx
        &< -\frac{\delta}{r^2} (\alpha^2 \|\hat{\vx}\|^2 + 2 \alpha \beta \ve^\top \hat{\vx} + \beta^2 \|\ve\|^2) \\
        &= -\frac{\delta}{r^2} \|\alpha \hat{\vx} + \beta \ve\|^2 \\
        &= - \delta
    ,\end{align*}
    contradicting our assumption that $\vzero$ is not a $(\delta,r)$-boundary QCSP.
\end{proof}

\begin{lemma}[Lemma~\ref{lem:approx_eigenvector}]
    Let $\delta, r > 0$, $\vx \in \R^f$ and $\eps \in (0,1)$.
    Let $f(\vy) = \frac 12 (\vy - \vx)^\top \vM (\vy - \vx)$ with $\lmax(\vM) \le L$.
    Let $\act$ be an affine subspace of $\R^d$ such that $\vx \in \act$.
    If $\min_{\vy \in \act \cap \ball(\vx, r)} f(\vy) < -\delta$,
        then Algorithm~\ref{alg:eigenvector} with $\step=\frac 1L$ with probability $1 - O(\xi)$ finds
        $\vy \in \act \cap \boundary{\ball(\vx, r)}$ with $f(\vy) \le -(1-\eps) \delta$
        after $T=O\left(\frac{L r^2}{\eps \delta} \log \left(\frac{L r d}{\xi \eps \delta}\right)\right)$ iterations\footnote{This is the only statement in this section where we consider a non-zero saddle point, due to its non-trivial role in Algorithm~\ref{alg:eigenvector}}.
\end{lemma}

\begin{proof}
    We show that Algorithm~\ref{alg:eigenvector} performs a power iteration on matrix $I - \step \vP \vM \vP$, where $\vP$ is the projection operator on linear subspace corresponding to $\act$.
    We show that $\lmin(\vP \vM \vP)$ corresponds to $\lmax(I - \step \vP \vM \vP)$.
    Finally, we use Fact~\ref{fact:power_convergence} to establish the convergence rate.

    By shifting the coordinate system so that $\vx$ becomes $\vzero$, we instead optimize function $g(\vy) = \frac 12 \vy^\top \vM \vy$
        on set $\ball(r) \cap \act_0$, where $\act_0 = \{\vy - \vx \mid \vy \in \act\}$ is the linear subspace parallel to $\act$.
    Let $\vP \in \R^{d \times d}$ be the projection operator on $\act_0$.
    Then for any $\vy \in \act_0$ we have $\vy = \vP \vy$ and for any $\vy \in \R^d$ we have $\proj_{\act_0}(\vy) = \vP \vy$.
    Defining $\vy_t = \vx_t - \vx$, we have:
    \begin{align*}
        \vy_{t+1}
        &= \vx_{t+1} - \vx
            \ifarxiv && \else \\& \quad\quad \fi \text{(Algorithm~\ref{alg:eigenvector}, Line~\ref{line:power_method})} \\
        &= \proj_\act (\vx_t - \step (\nabla f(\vx_t) - \nabla f(\vx))) - \vx 
             \ifarxiv && \else \\& \quad\quad \fi \quad\quad (\proj_\act (\vy) - \vx = \proj_{\act_0} (\vy - \vx)) \\
        &= \proj_{\act_0} (\vx_t - \vx - \step (\nabla f(\vx_t) - \nabla f(\vx)))
            \ifarxiv && \else \\& \quad\quad \fi \text{($\nabla f(\vx) = \vM \vx$\ and \ $\proj_{\act_0}(\vy) = \vP \vy$)} \\
        &= \vP (\vx_t - \vx - \step \vM (\vx_t - \vx))
            \ifarxiv && \else \\& \quad\quad \fi \text{(Definition of $\vy_t$)} \\
        &= \vy_t - \step \vP \vM \vy_t
            \ifarxiv && \else \\& \quad\quad \fi \text{($\vP \vy_t = \vy_t$ since, by induction, $\vy_t \in \act_0$)} \\
        &= \vy_t - \step \vP \vM \vP \vy_t   \\
        &= (I - \step \vP \vM \vP)^{t+1} \vy_0
    .\end{align*}
    Therefore, Algorithm~\ref{alg:eigenvector} performs a power iteration for matrix $\vB = I - \step \vP \vM \vP$.
    Since $\lmax(\vP \vM \vP) \le \lmax (\vM) \le L$ and $\step = \frac 1L$, all eigenvalues of $\vB$ lie in $[0, 2]$, and its largest eigenvalue is $1 - \step \lmin(\vP \vM \vP)$.
    
    Since there exists $\ve$ with $\|\ve\|=r$ and $h(\ve) < -\delta$, we have
    \[
        \lmin(\vP \vM \vP) \le - \frac{f(\ve)}{\nicefrac{\|\ve\|^2}{2}} < -\frac{2 \delta}{r^2},
    \]
    which implies $\lmax(\vB) = 1 + \frac{2 \delta}{Lr^2}$.
    Our goal is to find $\vy \in \boundary{\ball(r)}$ with $f(\vy) < -(1-\eps) \delta$, meaning
    \[
        \frac{\vy^\top (\vP \vM \vP) \vy}{\|\vy\|^2} < -(1-\eps) \frac{2 \delta}{r^2}
    ,\]
    which is equivalent to
    \[
        \frac{\vy^\top \vB \vy}{\|\vy\|^2} > 1 + (1-\eps) \frac{2 \delta}{L r^2}
    .\]
    By Fact~\ref{fact:power_convergence}, with probability $1 - O(\xi)$ we find such $\vy$ after the following number of iterations:
    \begin{align*}
        T
        &=O\left(\frac{\log d - \log (\xi \eps \frac{2 \delta}{L r^2})}
                {\log \frac{1 + \nicefrac{2 \delta}{L r^2}}{1 + (1-\eps) \cdot \nicefrac{2 \delta}{L r^2}}}\right) \\
        &=O\left(\frac{\log \left(\frac{L r d}{\xi \eps \delta}\right)}
                {\log\left(1 + \frac{\eps \delta}{L r^2}\right)}\right) \\
        &=O\left(\frac{L r^2}{\eps \delta} \log \left(\frac{L r d}{\xi \eps \delta}\right)\right)
    .
    \end{align*}
    Rescaling $\vy$ so that $\|\vy\|=r$ finishes the proof.
\end{proof}

\begin{theorem}[Theorem~\ref{thm:copositive_case}]
    Let $\delta, r > 0$.
    Let $f(\vx) = \frac 12 \vx^\top \vM \vx$ with $\lmax(\vM) \le L$ and
        let $S = \{\vx \mid \vA \vx \le \vzero\}$ be defined by $k$ linear inequality constraints.
    If $\vx$ is a ($\delta,r$)-QCSP, then Algorithm~\ref{alg:matrix_copositivity} with probability at least $1-\xi$ finds a point $\vx \in S \cap \ball(r)$ with $f(\vx) < -\Omega(\delta)$
        using $\tilde O\left(\frac{L r^2 k 2^k}{\delta}\right)$ deterministic gradient oracle calls.
\end{theorem}
\begin{proof}

    For $\actInd \subseteq [k]$, we define $\act_\actInd = \{\vx \mid \vA_i^\top \vx = 0, i \in \actInd\}$.
    By induction, we'll prove the following: for any $\actInd \in [k]$,
        if Algorithm~\ref{alg:matrix_copositivity} didn't find point $\vx$ with $f(\vx) < -(1 - \frac{|\actInd| + 1}{k}\eps) \delta$
        after executing Algorithm~\ref{alg:eigenvector} on all sets $\act_{\tilde{\actInd}}$
        such that $\actInd \subseteq \tilde{\actInd} \subseteq [k]$, then
    \[
        \min_{\vx \in \act_\actInd \cap S \cap \ball(r)} f(\vx) \ge -\left(1 - \frac{|\actInd|}{k}\eps\right)\delta
    .\]
    Basically, we have a subroutine which can return $1-\eps$ approximation, and, since this approximation factor may accumulate with each additional active constraint, we instead use approximation $1-\frac{\eps}{k}$ for every set of constraints.

    We fix $\actInd$ and assume that the statement holds for all $\tilde{\actInd}$ such that $\actInd \subsetneq \tilde{\actInd} \subseteq [k]$.
    By induction hypothesis, if for all such $\tilde{\actInd}$
        we didn't find point $\vx$ with $f(\vx) < -(1 - \frac{|\actInd| + 2}{k}\eps)\delta$, then
    \[
        \min_{\vx \in \act_{\tilde{\actInd}} \cap S \cap \ball(r)} f(\vx) \ge -(1 - \frac{|\actInd| + 1}{k}\eps)\delta
    \]
    for all such $\tilde{\actInd}$.
    If
    \[
        \min_{\vx \in \act_{\actInd} \cap S \cap \ball(r)} f(\vx) < -(1 - \frac{|\actInd|}{k}\eps)\delta
    ,\]
    then, since on the boundary we have
    \[
        \min_{\actInd \subsetneq \tilde{\actInd} \subseteq [k]} \min_{\vx \in \act_{\tilde \actInd} \cap S \cap \ball(r)} f(\vx) \ge -(1 - \frac{|\actInd| + 1}{k}\eps)\delta
    ,\]
    by Lemma~\ref{lemma:copositivity_approx_eigen} and Lemma~\ref{lem:approx_eigenvector} we can find $\vx \in \act_{\actInd} \cap S \cap \ball(r)$ with $f(\vx) < -(1 - \frac{|\actInd| + 1}{k}\eps)\delta$ using $\tilde O\left(\frac{L k r^2}{\eps \delta}\right)$ gradient computations.
    Hence, if the algorithm didn't find such $\vx$, then, as required:
    \[
        \min_{\vx \in \act_\actInd \cap S \cap \ball(r)} f(\vx) \ge -(1 - \frac{|\actInd|}{k}\eps)\delta
    .\]
    Since there $2^k$ possible subsets of constraints, the total number of gradient oracle calls is $\tilde O\left(\frac{L k r^2 2^k}{\eps \delta}\right)$.
\end{proof}

\section{Detailed Algorithms and Proof Outline}
\label{app:proof_outline}


\begin{algorithm}[t!]
    \caption{\mcAlg$(\vx, S, \delta)$: Escaping from a saddle point}
    \label{alg:escaping_general_full}
    \SetKwInOut{Input}{input}
	\SetKwInOut{Output}{output}
	\Indentp{0.1em}
    \Input{Saddle point $\vx$, feasible set $S = \{\vy \mid \vA \vy \le \vb \in \R^k\}$, $\delta$ from definition of $\delta$-SOSP}
    \Output{either $\vu \in S \cap \ball(\vx, r)$ with $f(\vu) < f(\vx) - \frac{\delta}{3}$, or reports that $\vx$ is a $\delta$-SOSP}
    \nonl\textbf{parameters:} $r=\trueR$ (Def.~\ref{def:sosp}), noise parameter $\xi$ \\
    \begin{algbox}
    Let $\vrsg(\vx, \tilde \sigma)$ be an algorithm that returns $\vg$ such that $\|\vg^{(0)} - \nabla f(\vx)\| \le \tilde \sigma$ w.h.p. \label{line:vrsg} \label{line:quad_approx_start} \\
    \nonl\linecomment{Compute $\vH$~-- approximate Hessian} \\
    $\theta \gets \Theta(\frac{r}{\sqrt{d}})$, \quad $\tilde \sigma \gets \Theta(\frac{\rho r^2}{d})$  \\
    $\vg^{(0)} \gets \vrsg(\vx, \tilde \sigma)$ \\
    \lFor{$\ve_1, \ldots, \ve_d$~-- the standard basis in $\R^d$}{
        $\vg_i \gets \vrsg(\vx + \theta \ve_i, \tilde \sigma)$
    }
    Let $\vH \in \R^{d \times d}$ whose $i$-th column is $\frac{\vg_i - \vg^{(0)}}{\theta}$ \label{line:hessian} \\
    \end{algbox}
    \begin{algbox}
    \nonl\linecomment{$f(\vx + \vh) \approx f(\vx) + \vh^\top \vv + \frac 12 \vh^\top \vM \vh$} \\
    Sample $\zeta \sim \mathcal{N}(\vzero, \xi \frac{\delta}{r\sqrt{d}} \cdot I)$, \quad $\gamma \sim U[-\xi, \xi]$ \\
    $\vv \gets \vrsg(\vx, \Theta(\nicefrac{\delta}{r})) + \zeta$ \label{line:linear_term} \\
    $\vM \gets \vH + \gamma I$ \label{line:quad_term} \label{line:quad_approx_end} \\
    \end{algbox}
    \For{$\actInd$~-- every subset of constraints}{
        $\act \gets \{\vy \mid \vA_i^\top \vx = b_i,\ i \in \actInd\}$, where $\vA_i$ is the $i$-th row of $\vA$ \hfill\linecomment{Optimize in $\act$} \\
        $\vp \gets \proj_\act(\vx)$ \hfill\linecomment{Ball center in $\act$}\\
        \lIf{$\|\vp - \vx\| > r$}{\textbf{continue}}
        \begin{algbox}
        \nonl\linecomment{Optimization of $g(\vy) = \frac 12 \vy^\top \s{\vM} \vy + \vy^\top \s{\vv}$
            in $\s{S} \cap \ball_{\dim \act}(\s{r})$} \\
        $\s{r} \gets \sqrt{r^2 - \|\vp - \vx\|^2}$ \label{line:r_proj} \label{line:proj_start} \\
        Let $\vO \in \R^{d \times \dim \act}$ be an orthonormal basis of $\act$ \\
        $\s{S} \gets \{\vy \mid \vA (\vp + \vO \vy) \le \vb\}$ \hfill\linecomment{$\vy \in \s{S}$ guarantees that $\vp + \vO \vy \in S$} \label{line:s_proj}\\
        $\s{\vM} \gets \vO^\top \vM \vO$ \label{line:m_proj} \\
        $\s{\vv} \gets \vO^\top (\vv + \vM (\vp - \vx))$ \label{line:v_proj} \label{line:proj_end} \\
        \end{algbox}
        $\vu \gets \textsc{FindInside}(\vx, \delta, \vp, \vO, (\s{\vM}, \s{\vv}), (\s{r}, \s{S}))$ \hfill\linecomment{Algorithm~\ref{alg:eigenvector_general_full}} \\
        \lIf{$\vu \not= \bot$}{\Return $\vu$}
    }
    \textbf{report} that $\vx$ is a $\delta$-SOSP 
\end{algorithm}

\begin{algorithm}[t!]
    \caption{\ifarxiv\else\\\fi\textsc{FindInside}$(\vx, \delta, \vp, \vO, (\s{\vM}, \s{\vv}), (\s{r}, \s{S}))$}
    \label{alg:eigenvector_general_full}
    \SetKwInOut{Input}{input}
    \SetKwInOut{Output}{output}
	\Indentp{0.1em}
    \Input{$\vx$~-- saddle point, $\delta$ from definition of $\delta$-SOSP, $\vp$~-- projection of $\vx$ on $\act$ (affine subspace corresponding to active constraints), $\vO$~-- orthonormal basis of $\act$, $g(\vy) = \frac 12 \vy^\top \s{\vM} \vy + \vy^\top \s{\vv}$~-- objective in $\R^{\dim \act}$, $\s{S} \cap \ball(\s{r})$~-- feasible set in $\R^{\dim \act}$}
    \Output{$\vp + \vO \vy$ such that $\vy \in \s{S} \cap \ball(\s{r})$ and $f(\vp + \vO \vy) < f(\vx) - \Omega(\delta)$, if it exists.}
    \vspace{2mm}
    \begin{algbox}
    \nonl\linecomment{\textbf{Case 1}: Large gradient}\\
    Let $\vy \gets \argmin_{\vy \in \s{S} \cap \ball(\s{r})} \vy^\top \s{\vv}$ \label{line:large_grad} \\
    \lIf{$f(\vp + \vO \vy) < f(\vx)  - \frac{\delta}{3}$}{
        \Return $\vp + \vO \vy$ \label{line:large_grad_check}
    }
    \end{algbox}
    \begin{algbox}
    \nonl\linecomment{\textbf{Case 2}: Solution is in $\interior{\ball(\s{r})}$} \\
    Let $\vy \gets -\vM^{-1} \vv$ \label{line:newton_step} \\
    \lIf{$\vy \in \s{S} \cap \ball(\s{r})$ and $f(\vp + \vO \vy) < f(\vx)  - \frac{\delta}{3}$}{
        \Return $\vp + \vO \vy$ \label{line:newton_check}
    }
    \end{algbox}
    \begin{algbox}
    \nonl\linecomment{\textbf{Case 3}: Solution is in $\boundary{\ball(\s{r})}$} \\
    \nonl\linecomment{Matrix diagonalization} \\
    Let $\vQ$ be an orthogonal matrix and $\vLambda=diag(\lambda_1, \ldots, \lambda_{\dim \act})$ be
        such that $\|\vQ^\top \vLambda \vQ - \s{\vM}\| \le \frac{\delta}{10 r^2}$ \label{line:diagonalization} \label{line:interior_start} \\
    \nonl \linecomment{Rotation of the linear term} \\
    Let $\tilde \vv \gets \vQ \s{\vv}$ and $\tilde v_i$ be its $i$-th coordinate \label{line:linear_term_rotation} \\
    \nonl\linecomment{$\tilde \vy$ from Line~\ref{line:eigen_solution} must have norm $r$} \\
    Let $\{\mu\}_j$ be approximate real solutions of $\sum_{i=1}^d \frac{\tilde v_i^2}{(\mu - \lambda_i)^2} = \s{r}^2$  \label{line:solution_norm} \\
    \For{each $\mu_j$} {
        Let $\tilde \vy$ be a vector such that its $i$-th coordinate is $\tilde y_i \gets \frac{\tilde v_i}{\mu_j - \lambda_i}$ \label{line:eigen_solution} \\
        \nonl \linecomment{Correction of approximate solutions} \\
        $\vy \gets \proj_{\s{S} \cap \ball(\s{r})}(\vQ^\top \tilde \vy)$ \\
        \lIf{$f(\vp + \vO \vy) < f(\vx) - \frac{\delta}{3}$}{
            \Return $\vp + \vO \vy$ \label{line:interior_end}
        }
    }
    \end{algbox}
    \nonl\Return $\bot$ \hfill\linecomment{No solution in affine subspace $\act$}
\end{algorithm}

We first reiterate our main results:
\begin{theorem}[Theorem~\ref{thm:general_case} restated]
    Let $S = \{\vx \mid \vA \vx \le \vb\}$ be a set defined by an intersection of $k$ linear inequality constraints.
    Let $f$ satisfy Assumptions~\ref{ass:lipschitz} and~\ref{ass:sgd} and let $\min_{\vx \in S} f(\vx) = f^\star$.
    Then there exists an algorithm which for $\delta > 0$ finds a $\delta$-SOSP in $\tilde O(\frac{f(\vx_0) - f^\star}{\delta}d^3 ( 2^{k} + \frac{\sigma^2}{\delta^{\nicefrac{4}{3}}}))$ time using $\tilde O(\frac{f(\vx_0) - f^\star}{\delta}(d + \frac{d^3 \sigma^2}{\delta^{\nicefrac{4}{3}}}))$ stochastic gradient oracle calls.
\end{theorem}

Recall that we aim to find a $\delta$-SOSP, i.e. $\vx \in S$ such that
\[
    f(\vy) \ge f(\vx) - \delta \text{ for all } \vy \in S \cap \ball(\vx, r), \text{ where }r=\trueR
.\]
If $\vx$ is not a $\delta$-SOSP, our algorithm finds find a point $\vy \in S \cap \ball(\vx, r)$ which significantly decreases function value: $f(\vy) < f(\vx) - \Omega(\delta)$.
Therefore, if $\vx_0$ is the initial point, our algorithm requires $O(\frac{f(\vx_0) - f^\star}{\delta})$ iterations.

\subsection{Algorithm~\ref{alg:escaping_general_full}: Quadratic approximation}

We first focus on the Lines~\ref{line:quad_approx_start}-\ref{line:quad_approx_end} of Algorithm~\ref{alg:escaping_general_full}.
The goal is to build a quadratic approximation of the objective with some additional properties (see below).
To simplify the presentation, w.l.o.g. by shifting the coordinate system, we assume that the current saddle point is $\vzero$ and consider the quadratic approximation of the function:
$
    f(\vx) = \frac 12 \vx^\top \nabla^2 f(\vzero) \vx + \vx^\top \nabla f(\vzero)
$.
Since $f$ is $\rho$-Lipschitz and $r=\trueR$, in $\ball(r)$ the quadratic approximation deviates from $f$ by at most $\frac{\delta}{6}$ (see derivation before Definition~\ref{def:sosp}).
For a small value $\xi$ (to be specified later), we instead analyze a noisy function
\[
    f'(\vx) = \frac 12 \vx^\top \vM \vx + \vx^\top \vv + f(\vzero)
,\]
where (all lines refer to Algorithm~\ref{alg:escaping_general_full}):
\begin{enumerate}
    \item
        $\vM = \vH + \gamma I$ (Line~\ref{line:quad_term}), where $\vH$ is an approximation of the Hessian (Line~\ref{line:hessian}) and $\gamma$ is small uniform noise, which guarantees that $\vM$ is non-degenerate with probability $1$.
    \item
        $\vv = \vg + \zeta$ (Line~\ref{line:linear_term}), where $\vg$ is an approximation of the gradient and $\zeta$ is small Gaussian noise, which guarantees that all coordinates of $\vv$ are sufficiently separated from $0$ w.h.p.
        Additionally, with probability $1$, linear systems of the form $(\vM - \mu I) \vx = \vv$ with $\rank (\vM - \mu I) < d$
            don't have any solutions, simplifying the analysis.\footnote{This also holds in the subspaces,  see proof of  Theorem~\ref{thm:app_general_case}).}
\end{enumerate}
We show that $f'$ is a good approximation of $f$, and first,
    we need to show that $\vH$ and $\vg$ are good approximations of the Hessian and the gradient respectively.
For the gradient, we use algorithm $\vrsg$ from Line~\ref{line:vrsg}, which w.h.p. estimates the gradient with precision $\tilde \sigma$ using $\tilde O(\frac{\sigma^2}{\tilde \sigma^2})$ stochastic gradient oracle calls (see Appendix~\ref{app:general}, Lemma~\ref{lem:vrsg}).
For the Hessian, since the $i$-th column of $\nabla^2 f(\vzero)$ is by definition $\lim_{\tau \to 0} \frac{\nabla f(\tau \ve_i) - \nabla f(\vzero)}{\tau}$, using a sufficiently small $\tau$ and approximating $\nabla f(\tau \ve_i)$ and $\nabla f(\vzero)$ using $\vrsg$, we find good approximation of $\nabla^2 f(\vx)$.
By selecting parameters as in Algorithm~\ref{alg:escaping_general_full} and by guaranteeing that noises $\zeta$ and $\gamma$ are bounded, we show the following result:
\begin{lemma}[Lemma~\ref{lem:stochastic-approximation} restated]
    \label{lem:stochastic-approximation-full}
    Let $f$ satisfy Assumptions~\ref{ass:lipschitz} and~\ref{ass:sgd}.
    Let $f'(\vx) = f(\vzero) + \vx^\top \vv + \frac 12 \vx^\top \vM \vx$, where $\vv$ and $\vM$ are as in Algorithm~\ref{alg:escaping_general_full}.
    For $\delta > 0$, $r = \trueR$ we have $\|f'(\vx) - f(\vx)\| < \frac{\delta}{2}$ for all $\vx \in \ball(r)$ w.h.p.
\end{lemma}

\paragraph{Reducing Case \texorpdfstring{$\vx^\star \in \boundary{S}$}{v* on the boundary of S} to Case \texorpdfstring{$\vx^\star \in \interior{S}$}{v* in Int S}.}
\label{ssec:projection}

We now elaborate on Lines~\ref{line:proj_start}-\ref{line:proj_end} of Algorithm~\ref{alg:escaping_general_full}.
The goal is, as in Section~\ref{sec:copositivity}, to reduce the case $\vx^\star \in \boundary{S}$ to the case $\vx^\star \in \interior{S}$.
If $\vx^\star \in \boundary{S}$, then there exist a non-empty set $\actInd$ of constraints active at $\vx^\star$.
Then we work in affine subspace $\act$ where constraints from $\actInd$ are active.
We parameterize $\act$:
    if $\vp = \proj_\act (\vx)$ and $\vO \in \R^{d \times \dim \act}$ is an orthonormal basis of $\act$,
    then any point in $\act$ can be represented as $\vp + \vO \vy$ for $\vy \in \R^{\dim \act}$.
Defining $g(\vy) = f'(\vp + \vO \vy)$, since
\begin{align*}
    g(\vy)
    &= \nicefrac 12 (\vp + \vO \vy)^\top \vM (\vp + \vO \vy) + (\vO \vy)^\top \vp
    \\ 
    &= \nicefrac  12 \cdot  \vy^\top (\vO^\top \vM \vO) \vy + \vy^\top (\vO^\top \vM \vp + \vO^\top \vv) + const
,\end{align*}
minimizing $f'$ in $\act \cap S \cap \ball(r)$ is equivalent to minimizing $g$ in $\s{S} \cap \ball(\s{r})$ (Lines~\ref{line:r_proj}-\ref{line:v_proj}), where:
\begin{enumerate}
    \item $\s{S}$ is a set of points $\vy \in \R^{\dim \act}$ such that $\vp + \vO \vy \in S$, namely $\s{S} = \{\vy \mid \vA_i (\vp + \vO \vy) \le b_i,\ i \notin \actInd\}$.
        Hence, $\s{S}$ is defined by linear inequalities, similarly to $S$.
    \item $\s{r}$ is a radius such that condition $\vy \in \ball(\s{r})$ is equivalent to $\vp + \vO \vy \in \ball(r)$.
        Since $\vp$ is the projection of $\vzero$ on $\act$, we have $\vO^\top \vp = \vzero$, and hence $\|\vp + \vO \vy\|^2 = \|\vp\|^2 + \|\vO \vy\|^2$.
        Since $\vO$ is an orthonormal basis of $\act$, $\|\vO \vy\| = \|\vy\|$, and hence $\s{r} = \sqrt{r^2 - \|\vp\|^2}$.
\end{enumerate}
For $\vy^\star$ such that $\vx^\star = \vp + \vO \vy^\star$, no constraints from $\s{S}$ are active, and hence $\vx^\star \in \interior{\s{S}}$.

\subsection{Algorithm~\ref{alg:eigenvector_general_full}: Escaping when \texorpdfstring{$\vy^\star \in \interior{\s{S}}$}{y* in Int S}}

In this section, we find minimizer $\vy^\star$ of function $g(\vy) = \frac 12 \vy^\top \s{\vM} \vy + \vy^\top \s{\vv} + C$ in $\s{S} \cap \ball(\s{r})$, while assuming that $\vy^\star \in \interior{\s{S}}$.
Since the solutions we find can be approximate, we have to guarantee that the objective is not too sensitive to the change of its argument.
It suffices to guarantee that $\|\vv\|$ is bounded, since for any $\vy \in \ball(\s{r})$ and perturbation $\vh$ there exists $\tau \in [0, 1]$ such that:
\begin{align*}
    |g(\vy) - g(\vy + \vh)|
    &= |(\nabla g(\vy + \tau \vh))^\top \vh| \\
    &\le (\|\nabla g(\vzero)\| + L \|\vy + \tau \vh\|) \|\s{\vh}\|\\
    &\le (\|\vv\| + L (\s{r} + \|\vh\|)) \|\vh\|
,\end{align*}
where we used that the objective is $L$-smooth and hence
\[
    \|\nabla g(\vy + \tau \vh)\| \le \|\nabla g(\vzero)\| + L \|\vy + \tau \vh\|
.\]
We consider the situation when $\|\s{\vv}\|$ is large as a separate case.
Otherwise, for $\vy^\star$, there are only two options: either $\vy^\star \in \interior{\ball(\s{r})}$ or $\vy^\star \in \boundary{\ball(\s{r})}$.
Algorithm~\ref{alg:eigenvector_general_full} handles these cases, as well as the case when $\|\vv\|$ is large, separately. 

\paragraph{Case 1: $\|\s{\vv}\|$ is large (Lines~\ref{line:large_grad}-\ref{line:large_grad_check}).}
If $\|\s{\vv}\|$ is large and we can find $\vy$ with small $\vy^\top \s{\vv}$, the linear term alone suffices to improve the objective.
We show that, if such $\vy$ doesn't exist, then $g(\vy^\star)$ requires $\vy^\star \in \boundary{\s{S}}$, which contradicts that $\vy^\star \in \interior{\s{S}}$.
Below we assume that $\|\s{\vv}\|$ is bounded.

\paragraph{Case 2: $\vy \in \interior{\ball(\s{r})}$ (Lines~\ref{line:newton_step}-\ref{line:newton_check}).}
In this case, $\vy^\star$ is an unconstrained critical point of $g$, and hence it must satisfy $\nabla g(\vy) = \vzero$, implying
\[
    \s{\vM} \vy + \s{\vv} = \vzero
\]
which gives the unique solution $\vy = - \s{\vM}^{-1} \s{\vv}$.
since $\vM$ is a perturbed matrix, so is $\s{\vM}$, and hence $\s{\vM}$ is non-degenerate with probability $1$.
It remains to verify that $\vy \in \ball(\s{r}) \cap \s{S}$ and $\vy$ decreases the objective by $\Omega(\delta)$.

\paragraph{Case 3: $\vy \in \boundary{\ball(\s{r})}$ (Lines~\ref{line:interior_start}-\ref{line:interior_end}).}
Since the only active constraint at $\vy^\star$ is $c(\vy) = \frac 12(\|\vy\|^2  - \s{r}^2) = 0$,
    by the KKT conditions, any critical point must satisfy $\nabla g(\vy) = \mu \nabla c(\vy)$ for some $\mu \in \R$,
    which is equivalent to $\s{\vM} \vy + \s{\vv} = \mu \vy$.
    Hence, for any fixed $\mu$, $\vy$ must be a solution of linear system $(\s{\vM} - \mu I) \vy = -\s{\vv}$.
    When $\s{\vM} - \mu I$ is degenerate (i.e. $\mu$ is an eigenvalue of $\s{\vM}$), due to perturbation of $\s{\vv}$, the system doesn't have any solution with probability $1$.
    It leaves us with the case when $\s{\vM} - \mu I$ is non-degenerate, when there exists a unique solution $\vy(\mu) := - (\s{\vM} - \mu I)^{-1} \s{\vv}$.
    
    \textbf{Diagonalization.} Dependence of $(\s{\vM} - \mu I)^{-1}$ on $\mu$ is non-trivial, but for a diagonal $\s{\vM}$, the inverse can be found explicitly.
    Hence, we perform diagonalization of $\s{\vM}$: we find orthogonal $\vQ$ and diagonal $\vLambda$
        such that $\|\s{\vM} - \vQ^\top \vLambda \vQ\| < \eps$ (Line~\ref{line:diagonalization}) in time $O(d^3 \log \nicefrac 1\eps)$~\citep{PC99}.
    Setting $\eps = O(\nicefrac{\delta}{\s{r}^2})$, we guarantee that the function changes by at most $O(\delta)$ in $\ball(\s{r})$.
    The function
    \[
        \vy(\mu) = - (\vQ^\top \vLambda \vQ - \mu I)^{-1} \s{\vv}
    \]
    can be written as
    \[
        \vQ \vy(\mu) = -(\vLambda - \mu I) \vQ \s{\vv}
    ,\]
    and hence we can instead work with rotated vectors $\tilde \vy(\mu) := \vQ \vy(\mu)$ and $\tilde \vv := \vQ \s{\vv}$ (Line~\ref{line:linear_term_rotation}).
    
    \textbf{Finding candidate $\mu$.} Since $\tilde \vy(\mu) = - (\s{\vM} - \mu I)^{-1} \tilde \vv$, for the $i$-th coordinate of $\tilde \vy(\mu)$ we have
        $\tilde y_i(\mu) = \frac{\tilde v_i}{\mu - \lambda_i}$ (Line~\ref{line:eigen_solution}).
    Since we are only interested in $\vy(\mu) \in \boundary{\ball(\s{r})}$ and $\vQ$ is an orthogonal matrix, we must have (Line~\ref{line:solution_norm}):
    \[
        \|\vy(\mu)\|^2 = \|\tilde \vy(\mu)\|^2 = \sum_{i=1}^d \tilde y_i(\mu)^2 = \s{r}^2
    ,\]
    and hence
    \[
        \sum_i \frac{\tilde v_i^2}{(\mu - \lambda_i)^2} = \s{r}^2
    .\]
    After multiplying the equation by $\prod_i (\mu - \lambda_i)^2$, we get an equation of the form $p(\mu) = 0$,
        where $p$ is a polynomial of degree $2d$.
    We find roots $\mu_1, \ldots, \mu_{2d}$ of the polynomial in time $O(d^2 \log d \cdot \log \log \nicefrac 1\eps)$~\citep{PAN1987591},
        where $\eps$ is the required root precision.
    For each $i$, we compute $\vy(\mu_i)$ and verify whether it lies in $\ball(\s{r}) \cap \s{S}$ and improves the objective by $-\Omega(\delta)$.
        
    \textbf{Precision.} Since the roots of the polynomial are approximate, when $\mu$ is close to $\lambda_i$,
        even a small perturbation of $\mu$ can strongly affect $y_i(\mu) = \frac{\tilde v_i}{\mu - \lambda_i}$.
    We solve this as follows: since $\|\tilde \vy(\mu)\| = \s{r}$,
        for each $i$ we must have $|\tilde y_i(\mu)|\le \s{r}$, implying $|\mu - \lambda_i| \ge \frac{|\tilde v_i|}{\s{r}}$.
    Therefore, $\mu$ must be sufficiently far from any $\lambda_i$, where the lower bound on the distance depends on $\s{r} \le \trueR$ and on $|\tilde v_i|$.
    By adding noise to $\vv$ (Line~\ref{line:linear_term} of Algorithm~\ref{alg:escaping_general_full}),
        we guarantee that the noise is preserved in $\tilde \vv$ so that each coordinate is sufficiently separated from $0$ w.h.p. This is formalized in Appendix~\ref{app:general}, Lemma~\ref{lem:solve_general}.

\section{Missing Proofs from Section~\ref{app:proof_outline}}
\label{app:general}

\begin{algorithm}
    \caption{$VRSG(\vx, \tilde \sigma)$: Variance Reduced Stochastic Gradient}
    \label{alg:vrsg}
	\SetKwInOut{Output}{output}
    \nonl\textbf{parameters:} stochastic gradient descent variance $\sigma$, error probability $\eps$ \\
    \Output{$\vg$ such that $\|\vg - \nabla f(\vx)\| < \tilde \sigma$ with probability at least $1 - \eps$} 
    \lIf{$\sigma = 0$}{\Return $\nabla f(\vx)$}
    $K \gets \frac{2 \sigma^2}{\tilde \sigma^2}$, \quad $M \gets O(\log \nicefrac{d}{\eps})$ \\
    \For{$i = 1, \ldots, M$}{
        Sample $\vg_{i,1}, \ldots, \vg_{i,K}$~-- independently sampled stochastic gradients of $f$ at $\vx$ \\
        $\vm_i \gets \frac 1K \sum_{j=1}^K \vg_{i,j}$
    }
    \Return the coordinate-wise median of $\vm_1, \vm_2, \ldots, \vm_M$
\end{algorithm}

\begin{lemma}
    \label{lem:vrsg}
    Let stochastic gradient oracle satisfy Assumption~\ref{ass:sgd}.
    Then for any $\tilde \sigma > 0$, Algorithm~\ref{alg:vrsg} returns a vector $\vg$ with $\|\vg - \nabla f(\vx)\| < \tilde \sigma$ with probability $1-O(\eps)$ using $O(1 + \frac{\sigma^2}{\tilde \sigma^2} \log \frac{d}{\eps})$ stochastic gradient oracle calls.
\end{lemma}
\begin{proof}
    Since the stochastic gradients are sampled independently and $\E \|\vg_{i,j} - \nabla f(\vx)\|^2 \le \sigma^2$ by Assumption~\ref{ass:sgd}, for each $i$ we have
    \[
        \E \|\vm_i - \nabla f(\vx)\|^2
        = \E \|\frac 1K \sum_{j=1}^k \vg_{i,j} - \nabla f(\vx)\|^2
        \le \frac{\sigma^2}{K} = \frac{\tilde \sigma^2}{2}
    .\]
    Applying Fact~\ref{fact:median_trick} to $\vm_1, \ldots, \vm_M$, we have $\|\vg - \nabla f(\vx)\| < \tilde \sigma$ with probability
    \[
        1 - d \cdot e^{-\Omega(M)} = 1 - d \cdot e^{-\Omega(\log \nicefrac{d}{\eps})} = 1 - O(\eps) \qedhere
    \]
\end{proof}

\begin{lemma}[Lemma~\ref{lem:stochastic-approximation}]
    Let $f$ satisfy Assumptions~\ref{ass:lipschitz} and~\ref{ass:sgd}.
    Let $f'(\vx + \vh) = f(\vx) + \vh^\top \vv + \frac 12 \vh^\top \vM \vh$, where $\vv$ and $\vM$ are as in Algorithm~\ref{alg:escaping_general}.
    For $\delta > 0$, $r = \trueR$ we have $\|f'(\vx + \vh) - f(\vx + \vh)\| < \frac{\delta}{2}$ for all $\vh \in \ball(r)$ w.h.p.
\end{lemma}
\begin{proof}
    Recall that $\vv$ and $\vM$ are noisy estimations of the gradient and the Hessian (below we reiterate how they are computed).
    By the Taylor expansion of $f$, we have
    \[
        f(\vx + \vh) = f(\vx) + \vh^\top \nabla f(\vx) + \frac 12 \vh^\top \nabla^2 f(\vx) \vh + \phi(\vh)
    ,\]
    where $\phi(\vh) = O(\|\vh\|^3)$.
    To bound $|f'(\vx + \vh) - f(\vx + \vh)|$, we analyze difference between $f$ and $f'$ for linear, quadratic and higher-order terms separately.
    Namely, we bound $|\vh^\top \nabla f(\vx) - \vh^\top \vv|$, $|\frac 12 \vh^\top \nabla^2 f(\vx) \vh - \frac 12 \vh^\top \vM \vh|$, and $|\phi(\vh)|$ with $\frac{\delta}{6}$ each.
    

    \paragraph{Higher-order terms}
    Since $f$ has a $\rho$-Lipschitz Hessian, we have $|\phi(\vh)| \le \frac{\rho r^3}{6} \le \frac{\delta}{6}$ for $r=\trueR$.
    
    \paragraph{Linear terms}
    Recall that $\vv = \vg + \zeta$, where $\vg = \vrsg(\vx, O(\nicefrac{\delta}{r}))$ and $\zeta \sim \mathcal{N}(\vzero, \xi \frac{\delta}{r\sqrt{d}} \cdot I)$ for some small $\xi$.
    By Lemma~\ref{lem:vrsg}, $\|\vg - \nabla f(\vx)\| < O(\nicefrac{\delta}{r})$ w.h.p.
    By choosing the appropriate constants, with probability $1 - O(\eps)$ we have
    \[
        \|\vv - \nabla f(\vx)\|
        = \|\vg - \nabla f(\vx)\| + \|\zeta\|
        \le \frac{\delta}{12r} + \xi \frac{\delta}{r} \log \frac{1}{\eps}
        \le \frac{\delta}{6r}
    ,\]
    where we selected $\xi \le \frac{1}{12} \log^{-1} \frac{1}{\eps}$.
    Using this bound, for all $\vh \in \ball(r)$ we have:
    \[|\vh^\top \nabla f(\vx) - \vh^\top \vv| \le r \|\vv - \nabla f(\vx)\| < \frac{\delta}{6}\]
    
    \paragraph{Quadratic terms}
    Recall that $\vM = \vH + \gamma I$, where $\gamma \sim U [-\xi, \xi]$ and $\vH$ is as in Algorithm~\ref{alg:escaping_general}:
        for $\{\ve_i\}$~-- the standard basis in $\R^d$, $\theta = \Theta(\frac{\delta}{\rho r^2})$ and $\tilde \sigma = \Theta(\frac{\delta^2}{\rho r^2})$,
        the $i$-th column of $\vH$ is $\frac{\vg_i - \vg^{(0)}}{\theta}$ with $\vg_i = \vrsg(\vx + \theta \ve_i, \tilde \sigma)$ and $\vg^{(0)} = \vrsg(\vx + \theta \ve_i, \tilde \sigma)$.
    For all $\vh \in \ball(r)$ we have:
    \[
        |\frac 12 \vh^\top \vM \vh - \frac 12 \vh^\top \nabla^2 f(\vx) \vh|
        \le |\frac 12 \vh^\top (\vH - \nabla^2 f(\vx)) \vh| + \frac{|\gamma|}{2} \|\vh\|^2
    \]
    The second term is bounded by selecting a sufficiently small $\gamma$ (i.e. a sufficiently small $\xi$), and hence it remains to bound the second term.
    Let $\vw_i$ be the difference between $\frac{\nabla f(\vx + \theta \ve_i) - \nabla f(\vx)}{\theta}$ and its stochastic estimate:
    \[
        \vw_i = \frac{(\vg_i - \vg^{(0)}) - (\nabla f(\vx + \theta \ve_i) - \nabla f(\vx))}{\theta}
    .\] 
    Since $\vg_i$ and $\vg^{(0)}$ are within distance $\tilde\sigma$ of the true gradients,
        we have $\|\vw_i\|=O\left(\frac{\tilde \sigma}{\theta}\right)$.
    Since $\nabla^2 f(\vx) \ve_i$ is the $i$-th column of $\nabla^2 f(\vx)$, we have:
    \begin{align*}
        &\frac{\vg_i - \vg^{(0)}}{\theta} - \nabla^2 f(\vx) \ve_i \\
        &= \vw_i + \frac{\nabla f(\vx + \theta \ve_i) - \nabla f(\vx)}{\theta} - \nabla^2 f(\vx) \ve_i \\
        &= \vw_i + \int_{\tau=0}^{1}(\nabla^2 f(\vx + \tau \theta \ve_i) - \nabla^2 f(\vx)) d\tau \cdot \ve_i \\
        &= \vw_i + \vu_i
    ,\end{align*}
    where $\|\vu_i\| = O(\rho \theta)$ since $\nabla^2 f$ is $\rho$-Lipschitz. Therefore, for any $\vh  \in \ball(r)$ we have
    \[
        (\vH - \nabla^2 f(\vx)) \vh = \sum_{i=1}^d h_i(\vw_i + \vu_i)
    ,\]
    and hence:
    \begin{align*}
        |\vh^\top (\vH - \nabla^2 f(\vx)) \vh|
        &= |\sum_{i=1}^d h_i (\vw_i + \vu_i)^\top \vh| \\
        &\le \sum_{i=1}^d |h_i| \cdot \|\vw_i + \vu_i\| \cdot \|\vh\| \\
        &= O\left(\|\vh\|_1 (\frac{\tilde \sigma}{\theta} + \rho \theta) \|\vh\|\right) \\
        &= O\left(\sqrt{d} r^2 ( \frac{\tilde \sigma}{\theta} + \rho \theta)\right),
    \end{align*}
    where the third line follows from $\|\vw_i\| = O(\frac{\tilde \sigma}{\theta})$ and $\|\vu_i\| = O(\rho \theta))$.
    Hence, it suffices to choose $\theta$ so that:
    \[
        \sqrt{d} r^2 \rho \theta = \Theta(\delta) \iff \theta = \Theta(\frac{\delta}{\sqrt{d} \rho r^2}) = \Theta(\frac{r}{\sqrt{d}})
    ,\]
    where we used $r = \trueR$.
    Similarly, we choose $\tilde \sigma$ so that:
    \[
        \sqrt{d} r^2\frac{\tilde \sigma}{\theta} = \Theta(\delta)
        \iff \tilde \sigma = \Theta(\frac{\delta \theta}{\sqrt{d} r^2}) =\Theta(\frac{\rho r^2}{d})
    .\]
\end{proof}

\begin{lemma}
    \label{lem:large_gradient}
    Let $\delta, r > 0$.
    Let $f(\vx) = \frac 12 \vx^\top \vM \vx + \vx^\top \vv + C$, with the following conditions satisfied:
    \begin{itemize}
        \item $\lambda_{\max}(\vM) \le L$
        \item (Large linear term) $\|\vv\| \ge 4 (L d r + \frac{(C + \delta) d}{r})$
        \item For $\vx^\star = \argmin_{\vx \in S \cap \ball(r)} f(\vx)$ we have $f(\vx^\star) < -\delta$
    \end{itemize}
    Then either $(\vx^\star)^\top \vv < - \frac{r}{8d} \|\vv\|$ or $\vx^\star \in \boundary{S}$.
\end{lemma}
\begin{remark}
    As we show in the proof of Lemma~\ref{lem:solve_general}, if $\|\vv\| \ge L d r + \frac{(C + \delta) d}{r}$ and $\vx^\star \notin \boundary{S}$, the linear term alone is sufficient to sufficiently improve the function, and the quadratic term is negligible.
\end{remark}

\begin{proof}
    For contradiction, assume that $(\vx^\star)^\top \vv \ge - \frac{r}{4 d} \|\vv\|$ and $\vx^\star \in \interior{S}$.
    We show that we can perturb $\vx^\star$ so that the point still lies in $\ball(r) \cap S$ and the objective decreases.
    We consider two cases: when $\vx^\star \in \interior{\ball(r)}$, we can simply move in the direction of $-\vv$.
    When $\vx^\star \in \boundary{\ball(r)}$, we rely on the fact that $(\vx^\star)^\top \vv < - \frac{r}{8d} \|\vv\|$, which guarantees that small change in direction $-\vv$ requires a small change in the orthogonal direction.

    \paragraph{Case 1: $\vx^\star \in \interior{\ball(r)}$.}
    Let $\vx^\star = \alpha \vv + \vw$, where $\vw \perp \vv$.
    Since $\vx^\star \in \interior{(\ball(r) \cap S)}$, for a sufficiently small $\eps$ we have $\vx^\star - \eps \vv \in \ball(r) \cap S$,
        and then:
    \begin{align*}
        &f(\vx^\star - \eps \vv) \\
        &= \frac 12 (\vx^\star - \eps \vv)^\top \vM (\vx^\star - \eps \vv) + (\vx^\star - \eps \vv)^\top \vv + C \\
        &= f(\vx^\star) - \eps (\vx^\star)^\top \vM \vv + \eps^2 \frac 12\vv^\top \vM \vv - \eps \|\vv\|^2 \\
        &\le f(\vx^\star) + \eps (r L \|\vv\| - \|\vv\|^2) + O(\eps^2) \\
        &< f(\vx^\star)
    ,\end{align*}
    contradicting the fact that $\vx^\star$ is the minimizer.
    
    \paragraph{Case 2: $\vx^\star \in \boundary{\ball(r)}$.}
    Let $\vx^\star = -\alpha \vv + \vw$, where $\vw \perp \vv$.
    Since $(\vx^\star)^\top \vv \ge - \frac{r}{8 d} \|\vv\|$, we have $\alpha \le \frac{r}{8 d \|\vv\|}$, and hence $\|\vw\| \ge \frac{r}{2}$ and $\alpha \|\vv\| \le \frac{\|w\|}{4d}$.
    We'll construct $\vx'$ such that $\vx' \in \ball(r)$, $\vx'$ is arbitrarily close to $\vx^\star$, and, since $\vx^\star \in \interior{S}$, we also have $\vx' \in \interior{S}$.
    Let $\vx' = -(\alpha + \eps) \vv + (1 - \theta) \vw$ for $\eps > 0$.
    To have $\vx' \in \ball(r)$, it suffices to guarantee that $\|\vx'\| \le \|\vx^\star\|$, meaning:
    \begin{align*}
        &(\alpha + \eps)^2 \|\vv\|^2 + (1 - \theta)^2 \|\vw\|^2
            \le \alpha^2 \|\vv\|^2 + \|\vw\|^2 \\
        &\iff (2 \alpha \eps + \alpha \eps^2) \|\vv\|^2
            \le \theta (2 - \theta) \|\vw\|^2
    ,\end{align*}
    and hence for $\vx^\star \in \boundary{\ball(r)}$ it suffices to select $\theta = \frac{2 \alpha \eps \|\vv\|^2}{\|\vw\|^2}$.
    Since $\theta = O(\eps)$, $\|\vx' - \vx^\star\|=O(\eps)$, and hence we can select $\vx'$ arbitrarily close to $\vx^\star$, which guarantees $\vx' \in S$.
    It remains to show that $f(\vx') < f(\vx^\star)$. Using $\theta = \frac{2 \alpha \eps \|\vv\|^2}{\|\vw\|^2} \le \frac{\eps}{2 d} \frac{\|\vv\|}{\|\vw\|}$:
    \begin{align*}
        f(\vx')
        &= f(-(\alpha + \eps) \vv + (1 - \theta) \vw) \\
        &= \frac 12 (-(\alpha + \eps) \vv + (1 - \theta) \vw)^\top \vM (-(\alpha + \eps) \vv \\
            & \quad\quad + (1 - \theta) \vw) + ((\alpha + \eps) \vv + (1 - \theta) \vw)^\top \vv + C \\
        &= f(\vx^\star) + \frac 12 (-\eps \vv - \theta \vw)^\top \vM (-\eps \vv - \theta \vw)  \\
            & \quad\quad + (-\eps \vv - \theta \vw)^\top \vM \vx^\star + (-\eps \vv - \theta \vw)^\top \vv \\
        &\le f(\vx^\star) + (\eps + \frac{\eps}{2 d})\|\vv\| L r + O(\eps^2) - \eps \|\vv\|^2 (\eps - \frac{\eps}{2 d}) \\
        &< f(\vx^\star)
    .\end{align*}
\end{proof}

\begin{lemma}
    \label{lem:solve_general}
    For $\delta, r, \theta > 0$, let $S = \{\vx \mid \vA \vy \ge \vb\}$ and let $f(\vx) = \frac 12 \vx^\top \vM \vx + \vx^\top \vv + C$ satisfy the following conditions:
    \begin{compactenum}
        \item $\lmax(\vM) \le L$,
        \item $(\vM - \mu I) \vy = -\vv$ has no solutions for all $\mu$ such that $\rank (\vLambda - \mu I) < d$,
        \item $|\tilde v_i| \ge \theta$ for all $i$, where $\tilde \vv = \vQ \vv$ and $\vQ$ is an orthogonal matrix defined in Line~\ref{line:diagonalization} of Algorithm~\ref{alg:eigenvector_general}.
    \end{compactenum}
    Let $\vx^\star= \argmin_{\vy \in S \cap \ball(r)} f(\vx)$ and assume that $f(\vx^\star) < -\delta$ and $\vx^\star \in \interior{S}$.
    Then Algorithm~\ref{alg:eigenvector_general} finds $\vx \in S \cap \ball(r)$ such that $f(\vx) < -(1 - \eps)\delta$ in time
    $\tilde O(d^3)$.
\end{lemma}
\begin{remark}
    In the statement above, we select $\theta$ sufficiently small so that adding random noise to $\nabla f(\vzero)$
    guarantees that condition 3 is satisfied w.h.p.
    Condition 2 is not required, but it's easy to satisfy and simplifies the analysis.
\end{remark} 
\begin{proof}
    As outlined in Section~\ref{sec:general-case}, we analyze the cases from Algorithm~\ref{alg:escaping_general} separately.
    We consider the situation when $\|\vv\|$ is large as a separate case, since in this case the function is sensitive to the change of its argument.
    Otherwise, for $\vx^\star$, there are only two options: either $\vx^\star \in \interior{\ball(r)}$ or $\vx^\star \in \boundary{\ball(r)}$.
    
    \paragraph{Case 1: Large $\|\vv\|$.} If $\|\vv\| \ge 4(L d r + \frac{(C+\delta) d}{r})$, then by Lemma~\ref{lem:large_gradient} either $(\vx^\star)^\top \vv < - \frac{r}{4d} \|\vv\|$ or $\vx^\star \in \boundary{S}$.
    Since we assume that $\vx^\star \in \interior{S}$, it leaves us with the case $(\vx^\star)^\top \vv < - \frac{r}{4d} \|\vv\|$.
    Then Algorithm~\ref{alg:escaping_general} finds $\vx$ with $\vx^\top \vv < - \frac{r}{d} \|\vv\|$ and hence
    \begin{align*}
        f(\vx)
        &\le \frac 12 Lr^2 + C - \frac{r}{4d} \|\vv\|^2 \\
        &\le \frac 12 Lr^2 + C - \frac{r}{4d} 4(Ldr + (C + \delta) \frac{d}{r}) \\
        &\le -\delta
    .\end{align*}

    \paragraph{Case 2: $\vx^\star \in \interior{S} \cap \interior{\ball(r)}$.}
    In this case, $\vx^\star$ is an unconstrained critical point, and so we must have $\nabla f(\vx) = \vM \vx + \vv = \vzero$.
    If $\vM$ is degenerate, by condition 2, there are no solutions.
    Otherwise, there exists a unique $\vx$ such that $\vM \vx = -\vv$.
    We then check that $\vx \in S \cap \ball(r)$ and $f(\vx) < f(\vzero) - \frac{\delta}{2}$.

    \paragraph{Case 3: $\vx^\star \in \interior{S} \cap \boundary{\ball(r)}$.} By the method of Lagrangian multipliers,
        we must have $\vM \vx + \vv = \mu \vx$ for some $\lambda$.
    By condition 2, if $\rank(\vM - \mu I) < d$, then $(\vM - \mu I) \vx = -\vv$ has no solutions.
    Hence, it suffices to consider cases when $\mu$ is not an eigenvalue of $\vM$.

    When $\mu$ is not an eigenvalue of $\vM$, $(\vM - \mu I) \vx = -\vv$ has a unique solution.
    Based on~\citep{PC99}, we find orthogonal $\vQ$ and diagonal $\vLambda$
        such that $\|\vM - \vQ^\top \vLambda \vQ\| < \eps$,
        which can be done in time $O(d^3 \log \nicefrac 1\eps)$.
    By setting $\eps = O(\nicefrac{\delta}{r^2})$, we guarantee that the function changes by at most $O(\delta)$ in $\ball(r)$.
    Then $(\vQ^\top \vLambda \vQ - \mu I) \vx = -\vv$ is equivalent to
    \[
        (\vLambda - \mu I) \vQ \vx = - \vQ \vv
    .\]
    Let $\tilde \vv = \vQ \vv$ and $\tilde \vx = \vQ \vx$.
    Then the solutions for $\tilde \vx$ have coordinates $\tilde x_i = \frac{-\tilde v_i}{\mu - \lambda_i}$ for all $i$.
    Since we are only interested in solutions on $\boundary{\ball(r)}$ and $\vQ$ is an orthogonal matrix, we must have
    \[
        \|\vx\|^2 = \|\tilde \vx\|^2 = \sum_{i=1}^d \frac{\tilde v_i^2}{(\mu - \lambda_i)^2} = r^2
    .\]
    By multiplying both sides by $\prod_i (\mu - \lambda_i)^2$, we get an equation of form $p(\lambda) = 0$,
    where $p$ is the polynomial with degree $2d$ and the highest degree coefficient being $r^2$.
    We can find all the roots of $p$ with precision $\nu$ (to be specified later)
        in time $O(d^2 \log d \cdot \log \log \frac{1}{\nu})$~\citep{PAN1987591}.
    Since we are only interested in real roots, we take the real parts of the approximate roots (if the exact root is real, taking the real part only improves the accuracy).

    Next, we estimate the required precision $\nu$.
    Since we have $\|\tilde \vx\| = r$, it requires, in particular, that $|x_i| \le r$ for all $i$, which means that
    $\lambda_i$ must satisfy
    \[
        \frac{|\tilde v_i|}{|\mu - \lambda_i|} \le r \iff |\mu - \lambda_i| \ge \frac{|\tilde v_i|}{r}
    .\]
    Since $|v_i| \ge \theta$ by our assumption, we have $|\mu - \lambda_i| \ge \frac{\theta}{r}$.

    Let $\mu$ be the exact root of the polynomial and $\mu' \in [\mu - \nu, \mu + \nu]$
        be the corresponding approximate root found the algorithm.
    Let $\tilde \vx$ and $\tilde \vx'$ be the points in $\R^d$ corresponding to $\mu$ and $\mu'$.
    Then we have:
    \begin{align*}
        |\tilde x_i - \tilde x'_i|
        &= |\frac{\tilde v_i}{\mu - \lambda_i} - \frac{\tilde v_i}{\mu' - \lambda_i}| \\
        &= |\frac{\tilde v_i}{\mu - \lambda_i}\cdot \frac{\mu - \mu'}{\mu' - \lambda_i} | \\
        &\le \frac{r \nu}{|\mu' - \lambda_i|} \\
        &\le \frac{r \nu}{\frac{\theta}{r} - \nu} \\
        &\le \frac{2 r^2 \nu}{\theta}
    ,\end{align*}
    where we assumed $\nu < \frac{\theta}{2 r}$.
    If $\vx = \vQ^\top \tilde \vx$ and $\vx' = \vQ^\top \tilde \vx$, then
        $\|\vx - \vx'\| = \|\tilde \vx - \tilde \vx'\| \le \frac{2 \sqrt{d} r^2 \nu}{\theta}$.

    Let $\vx''$ be the projection of $\vx'$ on $\ball(r) \cap S$.
    Then $\|\vx'' - \vx\| \le 2 \|\vx' - \vx\|$, and hence:
    \begin{align*}
        |f(\vx) - f(\vx'')|
        &= \left|\frac 12 (\vx^\top \vM \vx - \vx''^\top \vM \vx'') + (\vx - \vx'')^\top \vv\right| \\
        &= \left|\frac 12 (\vx - \vx'')^\top \vM (\vx + \vx'') + (\vx - \vx'')^\top \vv\right| \\
        &\le \|\vx - \vx''\| (L r + \|\vv\|) \\
        &\le \nu \cdot \frac{4 \sqrt{d} r^2}{\theta} (L (d + 1) r + \frac{(C+\delta)d}{r}) \\
        &\le \frac{\delta}{10},
    \end{align*}
    by selecting $\nu < \frac{\delta \theta}{40 \sqrt{d} (L (d + 2) r^3 + (C+\delta)dr)}$.
\end{proof}

\begin{theorem}[Theorem~\ref{thm:general_case}]
    \label{thm:app_general_case}
    Let $S = \{\vx \mid \vA \vx \le \vb\}$ be a set defined by an intersection of $k$ linear inequality constraints.
    Let $f$ satisfy Assumptions~\ref{ass:lipschitz} and~\ref{ass:sgd} and let $\min_{\vx \in S} f(\vx) = f^\star$.
    Then there exists an algorithm which for $\delta > 0$ finds a $\delta$-SOSP in $\tilde O(\frac{f(\vx_0) - f^\star}{\delta}d^3 (2^{k} + \frac{\sigma^2}{\delta^{\nicefrac{4}{3}}}))$ time using $\tilde O(\frac{f(\vx_0) - f^\star}{\delta}(d + \frac{d^3 \sigma^2}{\delta^{\nicefrac{4}{3}}}))$ stochastic gradient oracle calls w.h.p.
\end{theorem}
\begin{proof}
    It suffices to show that if $\vx$ is not a $\delta$-SOSP, then Algorithm~\ref{alg:escaping_general} decreases function value by $\Omega(\delta)$, and hence there can't be more that $O(\frac{f(\vx_0) - f^\star}{\delta})$ iterations. 
    W.l.o.g. we can assume that the current iterate is $\vzero$ and $f(\vzero)=0$.
    As shown in Lemma~\ref{lem:stochastic-approximation}, we can replace $f$
        with its quadratic approximation $\frac 12 \vx^\top \vM \vx + \vx^\top \vv$ while increasing the function value by at most $\frac{\delta}{2}$.
    
    As shown in Section~\ref{sec:general-case}, we can consider function $g(\vy) = \frac 12 \vy^\top \s{\vM} \vy + \vy^\top \s{\vv} + C$ in $\s{S} \cap \ball(\s{r})$.
    By Lemma~\ref{lem:solve_general}, we can find $\vy$ with $g(\vy) < -\frac 12 \delta$ in $\tilde O(d^3)$ time.
    It remains to show that the conditions from Lemma~\ref{lem:solve_general} are satisfied.
    Recall that for $\vO \in \R^{d \times \dim \act}$~-- an orthonormal basis of $\act$, we define the following:
    \begin{compactitem}
        \item $\s{S} \gets \{\vy \mid \vA (\vp + \vO \vy) \le \vb\}$
        \item $\s{\vM} \gets \vO^\top \vM \vO$
        \item $\s{\vv} \gets \vO^\top (\vv + \vM (\vp - \vx))$
    \end{compactitem}

    As required in Lemma~\ref{lem:solve_general}, $\s{S}$ is the set defined by linear inequality constraints.
    We next check other requirements in Lemma~\ref{lem:solve_general}.
    \paragraph{1) $\lmax(\s{\vM}) \le L$:} Since $\vO$ defines an orthonormal basis, $\lmax(\s{\vM}) = \lmax(\vO^\top \vM \vO) \le \lmax(\vM) \le L$.
    \paragraph{2) $(\s{\vM} - \mu I) \vy = -\s{\vv}$ has no solutions for all $\mu$ such that $\rank (\vLambda - \mu I) < d$}
    \[
        \s{\vv}
        = \vO^\top (\vv + \vM (\vp - \vx))
        = \vO^\top \zeta + \vO^\top (\vg + \vM (\vp - \vx))
    ,\]
    where $\zeta \sim \mathcal{N}(\vzero, \xi \frac{\delta}{r\sqrt{d}} \cdot I)$.
    Since $\vO$ is an orthonormal basis, we have $\vO^\top \zeta \sim \mathcal{N}(\vzero, \xi \frac{\delta}{r\sqrt{d}} \cdot I)$.
    This Gaussian noise guarantees that the $\s{\vv}$ has probability $0$ to lie in any fixed subspace, which guarantees that there are no solutions for degenerate systems of form $(\s{\vM} - \mu I) \vy = -\s{\vv}$.
    
    \paragraph{3) $|\tilde v_i| \ge \theta$ for all $i$}
    Recall that $\tilde \vv = \vQ \s{\vv}$ as defined in Line~\ref{line:linear_term_rotation} of Algorithm~\ref{alg:eigenvector_general},
        where $\vQ$ is an orthogonal matrix.
    For a fixed $i$ and $\theta$, we consider $\Pr_\zeta[|\tilde v_i| < \theta]$.
    Due to rotation invariance of Gaussian noise, its projection on any linear subspace
        has distribution $\mathcal N(\vzero, \xi \frac{\delta}{r \sqrt{d}} I)$.
    By selecting $\theta=\eps \frac{\xi \delta}{d^2 r}$, we have that the for all $i$, $|\tilde v_i| > \theta$ with
    probability at least $1 - \eps$.
    
    By selecting $\eps$ such that taking the union bound over all $O(\frac{f(\vx_0) - f^\star}{\delta})$ iterations, $d$ coordinates and $2^k$ sets of constraints gives bounded error probability, we finish the proof.
\end{proof}

\section{Experiments}
\label{app:experiments}

Before describing our experimental setup, we would like to discuss the choice of parameters in Algorithm~\ref{alg:escaping_general_full}:
\begin{itemize}
    \item $\delta$ is a parameter of the problem, and hence can only be determined by the user.
    \item Radius $r$ can be selected adaptively.
        I.e. if Algorithm~\ref{alg:eigenvector_general_full} finds a candidate which, according to the quadratic estimate, should decrease the objective value, but it in fact does not decrease, then the radius should be reduced.
    \item In practice, noise parameter $\xi$ can be set to $0$ or to a very small constant.
    \item Some parameters rely on $\sigma^2$, which can be estimated by the empirical variance of stochastic gradient.
\end{itemize}

For our experiments, we consider the quadratic objective $f(\vx) = \frac 12 \vx^\top \vA \vx$ under box constraints $x_i \in [-1, 1]$.
Such an objective appears in relaxations of graph problems: for example, when $\vA$ is an adjacency matrix, this objective can represent a graph partitioning problem with the goal of minimizing the cut between parts.
Optional additional constraints allow one to model the variations of the problem: for example, adding constraint $\sum_{i} x_i = 0$ leads to the relaxation of the balanced graph partitioning problem.

In our settings, $\vA$ is the adjacency matrix of an Erd\"os-R\'enyi graph with parameters $n=300$ and $p=0.05$.
Our choice of parameters is as follows.
\begin{itemize}
    \item We choose a sufficiently small $\delta = 10^{-2}$.
    \item While for purely quadratic objective any $\rho \ge 0$ is valid, and hence the radius $r$ can be infinite, for demonstration purposes we set $r = 10^{-1}$.
\end{itemize}

\begin{figure}[t!]
    \centering
    \begin{subfigure}[t]{\ifarxiv 0.45\columnwidth \else \columnwidth \fi}
        \centering
        \includegraphics[width=\columnwidth]{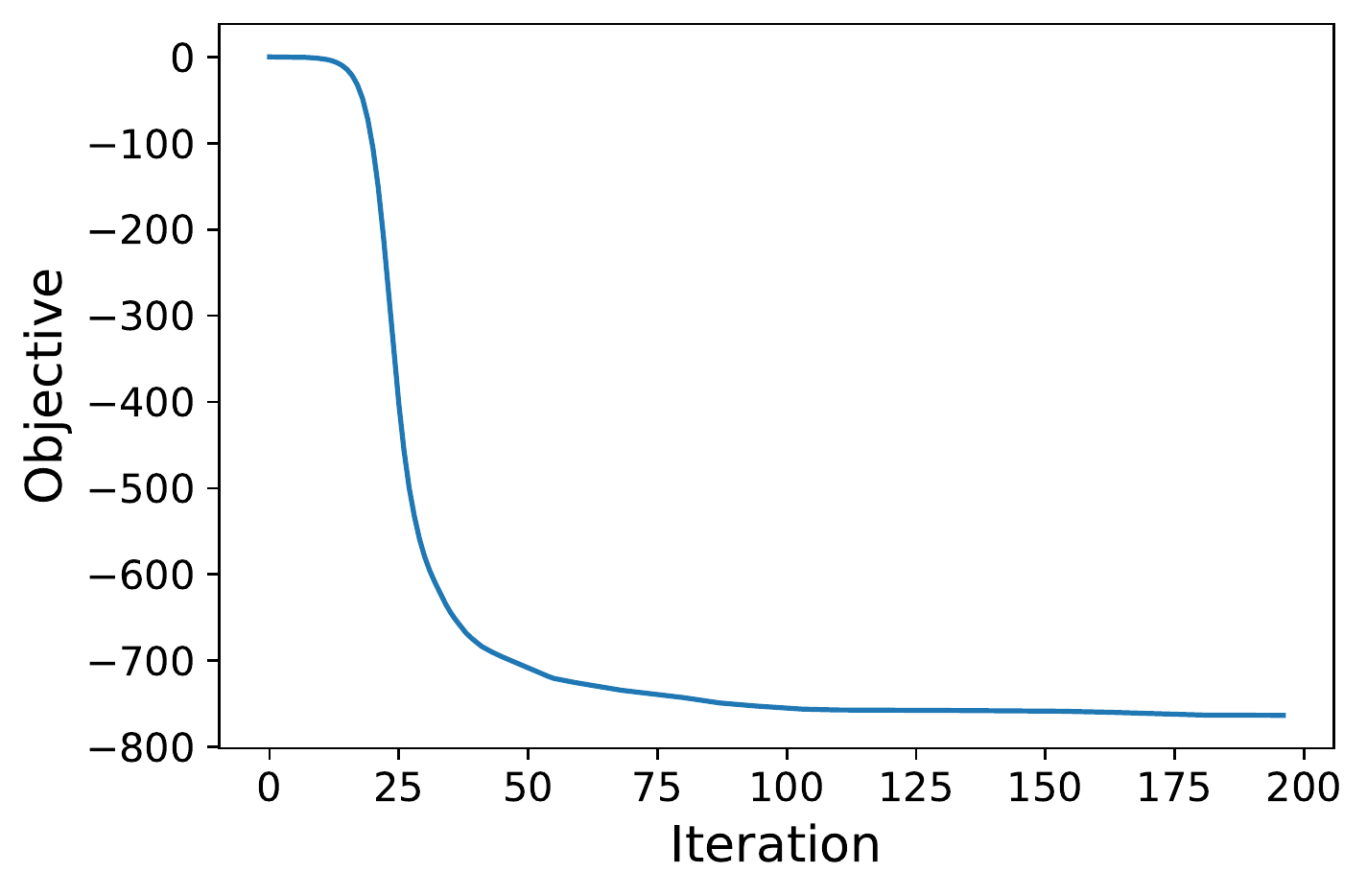}
        \caption{Objective value at every iteration}
    \end{subfigure}
    \begin{subfigure}[t]{\ifarxiv 0.45\columnwidth \else \columnwidth \fi}
        \centering
        \includegraphics[width=\columnwidth]{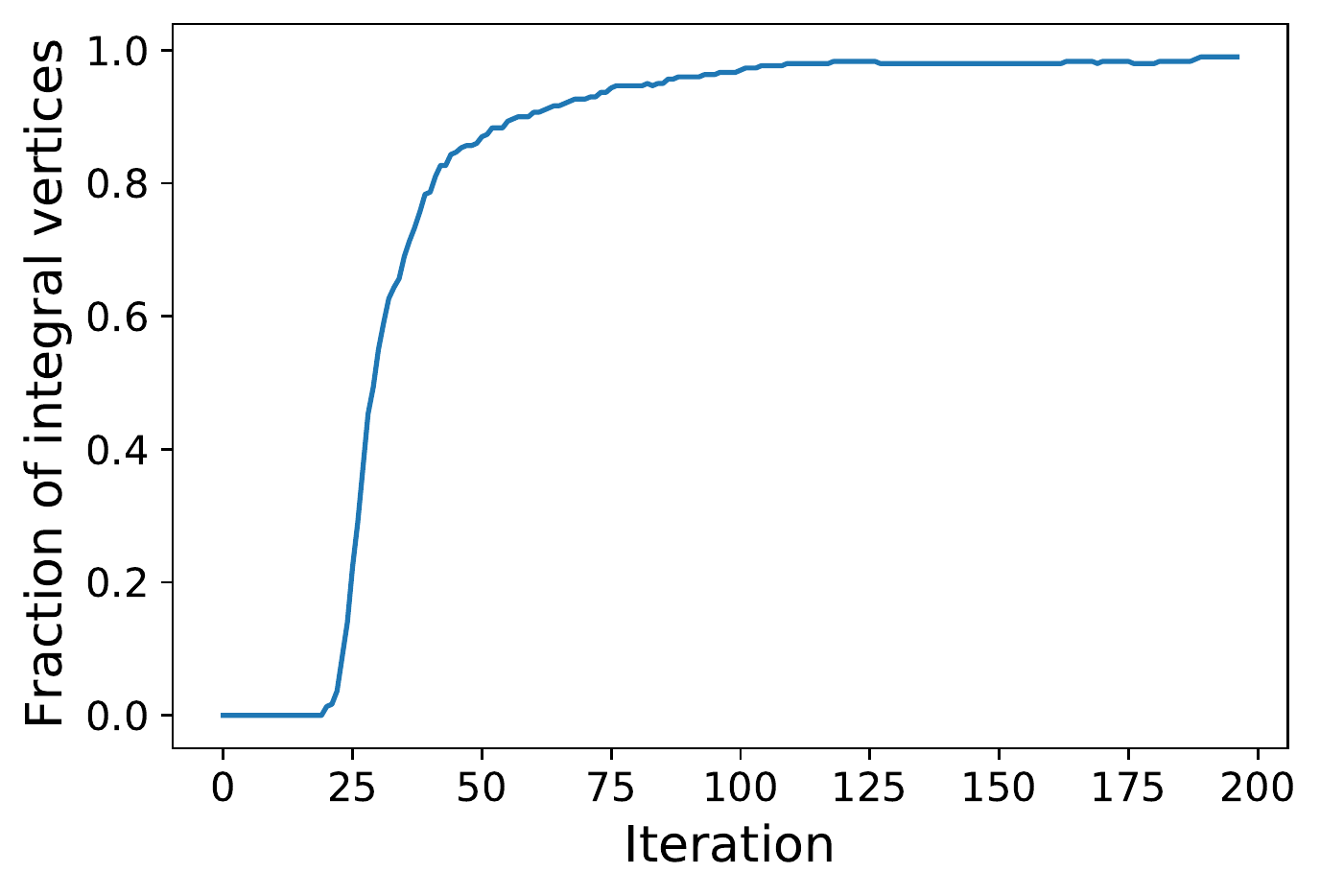}
        \caption{The fraction of vertices with values being $-1$ and $1$ at every iteration}
    \end{subfigure}
    \caption{Performance Algorithm~\ref{alg:escaping_general_full} on $f(\vx) = \frac 12 \vx^\top \vA \vx$ under constraints $\vx \in [-1,1]^n$, where $\vA$ is the adjacency matrix of an Erd\"os-R\'enyi graph with parameters $n=300$ and $p=0.05$}
    \label{fig:experiments}
\end{figure}

We show the performance of Algorithm~\ref{alg:escaping_general_full} on this objective function in Figure~\ref{fig:experiments}.
The objective rapidly decreases until the fraction of integral coordinates (i.e. with values either $-1$ or $1$) becomes close to $1$.
Notably, even though the theoretical running time is exponential in $n$, in practice we don't encounter this issue until almost all vertices become integral.

\end{document}